\documentclass[11pt]{article}

\usepackage[table]{xcolor} 
\usepackage{amsmath} 
\usepackage{booktabs}

\definecolor{LLightGray}{gray}{0.95} 
\definecolor{rowgray}{gray}{0.90}    

\usepackage[final]{acl}
\usepackage{booktabs}
\usepackage{times}
\usepackage{latexsym}
\usepackage{multirow}
\usepackage{makecell}
\usepackage[T1]{fontenc}

\usepackage[utf8]{inputenc}

\usepackage{microtype}

\usepackage{inconsolata}

\usepackage{graphicx}
\usepackage{xcolor}

\usepackage{amsmath}
\usepackage{amsfonts}
\usepackage{enumitem}

\usepackage{pifont}
\newcommand{\cmark}{\checkmark}
\newcommand{\xmark}{\ding{55}}

\usepackage{algorithm}
\usepackage{algpseudocode}

\definecolor{darkgreen}{RGB}{34,139,34}

\title{\textbf{Doc-$V^*$}: Coarse-to-Fine Interactive Visual Reasoning for Multi-Page Document VQA}

\author{
 \textbf{Yuanlei Zheng\textsuperscript{1}\thanks{Equal contribution}},
 \textbf{Pei Fu\textsuperscript{2}\footnotemark[\value{footnote}]},
 \textbf{Hang Li\textsuperscript{2}},
 \textbf{Ziyang Wang\textsuperscript{1}},
\\
 \textbf{Yuyi Zhang\textsuperscript{1}},
 \textbf{Wenyu Ruan \textsuperscript{1}},
 \textbf{Xiaojin Zhang\textsuperscript{3}},
 \textbf{Zhongyu Wei\textsuperscript{4}},
\\
 \textbf{Zhenbo Luo\textsuperscript{2}\thanks{Corresponding author}},
 \textbf{Jian Luan\textsuperscript{2}},
 \textbf{Wei Chen\textsuperscript{1}\footnotemark[\value{footnote}]},
 \textbf{Xiang Bai\textsuperscript{1}}
\\
 \textsuperscript{1}School of Software Engineering, Huazhong University of Science and Technology, \\
 \textsuperscript{2}MiLM Plus, Xiaomi Inc., \\
 \textsuperscript{3}School of Computer Science and Technology, Huazhong University of Science and Technology, \\
 \textsuperscript{4}School of Data Science, Fudan University
}

\begin{document}
\maketitle

\begin{abstract}
Multi-page Document Visual Question Answering requires reasoning over semantics, layouts, and visual elements in long, visually dense documents. Existing OCR-free methods face a trade-off between capacity and precision: end-to-end models scale poorly with document length, while visual retrieval-based pipelines are brittle and passive. We propose \textbf{Doc-$V^*$}, an \textbf{OCR-free agentic} framework that casts multi-page DocVQA as sequential evidence aggregation. \textbf{Doc-$V^*$} begins with a thumbnail overview, then actively navigates via semantic retrieval and targeted page fetching, and aggregates evidence in a structured working memory for grounded reasoning. Trained by imitation learning from expert trajectories and further optimized with Group Relative Policy Optimization, \textbf{Doc-$V^*$} balances answer accuracy with evidence-seeking efficiency. Across five benchmarks, \textbf{Doc-$V^*$} outperforms open-source baselines and approaches proprietary models, improving out-of-domain performance by up to \textbf{47.9\%} over RAG baseline. Other results reveal effective evidence aggregation with selective attention, not increased input pages. \href{https://github.com/SeerRay-Lab/Doc-V}{https://github.com/SeerRay-Lab/Doc-V}
\end{abstract}

\section{Introduction}
Understanding multi-page, visually rich documents—such as academic papers, financial reports, and industrial manuals—remains a core challenge in \textit{Document Visual Question Answering} (DocVQA)~\cite{docvqa,tito2023hierarchical}. Unlike plain text, such documents convey information through a complex interplay of textual semantics, spatial layouts, and visual elements (e.g., tables and figures)~\cite{deepl_survey}. Conventional \textbf{OCR-based} pipelines linearize document images into text before reasoning~\cite{OCR_servey,mineru,docformer}, but inevitably lose fine-grained layout cues and suffer from cascading OCR errors. Recent \textbf{OCR-free} or \textbf{pure-vision} approaches instead model documents directly as images using multimodal large language models (MLLMs)~\cite{lee2023pix2struct,donut,liu2024textmonkey}, enabling joint visual–semantic reasoning and improved robustness.

\begin{figure*}[t]
    \centering
    \includegraphics[width=0.95\linewidth]{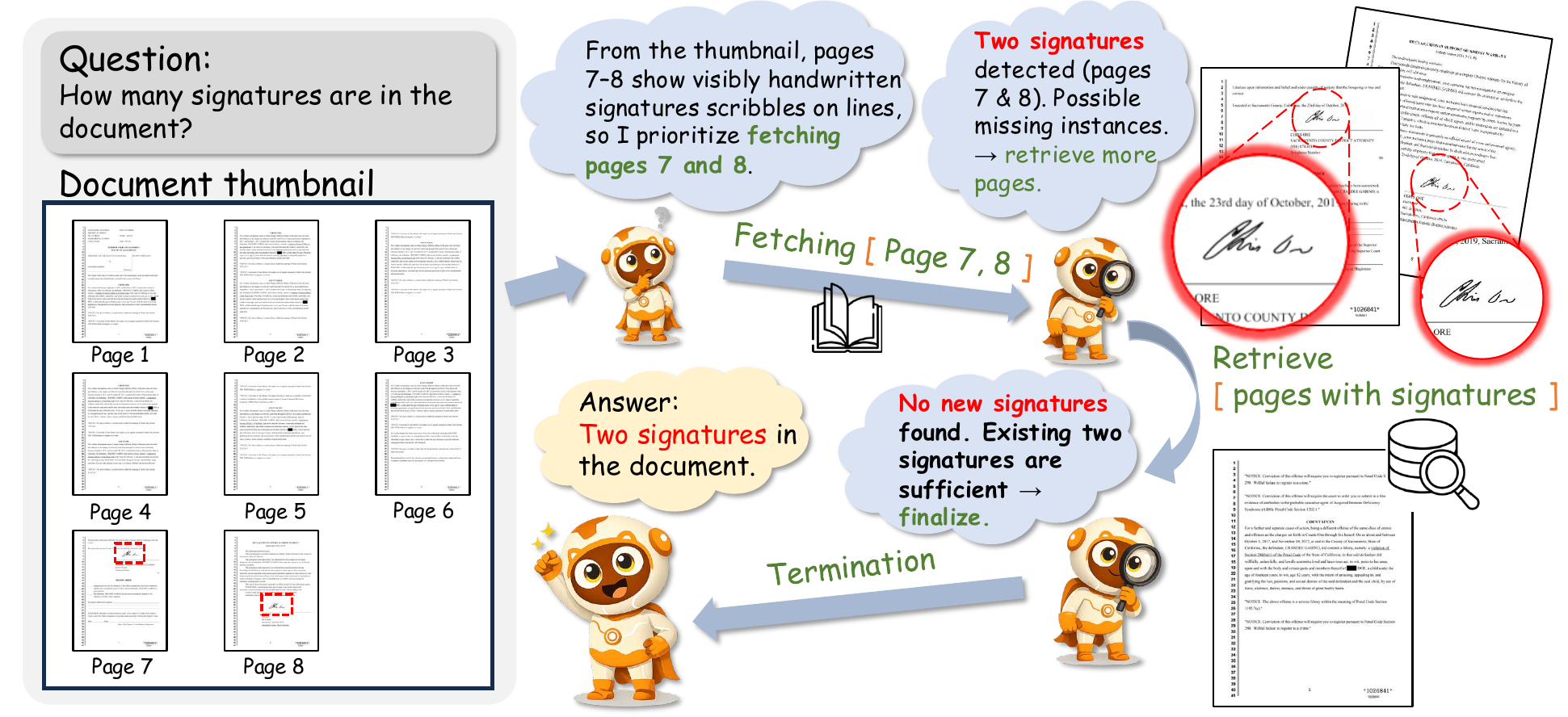}
    \caption{\textbf{The \textbf{Doc-$V^*$} agent workflow for multi-page document VQA.} It adopts an \textit{active perception} paradigm by planning from a global thumbnail view and iteratively deciding when to fetch high-resolution pages or perform semantic searches, aggregating evidence in a structured working memory for grounded answering.}
    \label{fig:teaser}
\end{figure*}

However, existing pure-vision methods face a fundamental trade-off between \textit{capacity} and \textit{precision}. \textbf{End-to-end} models process entire documents as long image sequences~\cite{zhu2025internvl3,hu2025mplug2,bai2025qwen25}, but scale poorly to long documents due to quadratic attention cost, context length limits, and the "\textit{lost-in-the-middle}" effect~\cite{liu2024lost}. In contrast, visual retrieval-augmented generation (\textbf{RAG}) systems reduce noise by retrieving top-$k$ pages before generation~\cite{cho2024m3docrag,faysse2025colpaliefficientdocumentretrieval,Urag}, yet suffer from retrieval errors, sensitivity to hyperparameters, and limited multi-hop reasoning. Critically, both paradigms remain \emph{passive}: they process a fixed input without adapting their strategy as new evidence emerges. 

We argue that this limitation arises from a mismatch with human document-reading behavior. Guided by \textit{Active Vision Theory}~\cite{aloimonos1988active}, human experts treat perception as a goal-directed process: they first obtain a global structural overview, then iteratively seek, verify, and integrate evidence while maintaining working memory. Inspired by this cognitive process, we propose \textbf{Doc-$V^*$}, an \textbf{OCR-free agentic framework} that formulates multi-page DocVQA as a \textbf{\emph{sequential evidence aggregation process}}. \textbf{Doc-$V^*$} begins with a \textit{Global Thumbnail Overview} that provides a low-cost structural prior, and then alternates between \textit{structured visual reasoning} and \textit{document navigation actions}, including semantic retrieval and targeted page fetching. This interactive reasoning allows the agent to \textbf{active perception} and \textbf{piece together discontinuous visual evidence} before answering. Figure~\ref{fig:teaser} shows the agent workflow of \textbf{Doc-$V^*$}.

To train \textbf{Doc-$V^*$}, we adopt a two-stage optimization strategy. We first perform supervised fine-tuning using high-quality interaction trajectories synthesized by GPT-4o, providing a strong cold start. We then apply Group Relative Policy Optimization (GRPO)~\cite{guo2025deepseek} to jointly optimize answer accuracy and evidence-seeking efficiency through reward signals that account for answer quality, evidence discovery, and format compliance. Extensive experiments on five benchmarks demonstrate that \textbf{Doc-$V^*$} consistently outperforms existing open-source baselines and rivals proprietary models like \textbf{GPT-4o}, particularly in out-of-domain settings where it achieves up to a \textbf{47.9\% improvement} over static RAG baselines, as well as \textbf{robustness} under variations in retrieval tools and hyperparameters. We also demonstrate that long-document understanding hinges on \textbf{effective aggregation of evidence} with selective attention rather than \textbf{sheer input pages}, which is crucial to the success of \textbf{Doc-$V^*$}. 

\section{Related Work}

Visual Document Question Answering (DocVQA) has progressed from single-page inputs to long and multi-page documents, driven by the increasing demand for handling complex real-world document understanding scenarios. Existing methods mainly follow two paradigms: 1) \textbf{OCR-based DocVQA} \quad OCR-based approaches first extract textual and layout structures via OCR and document parsing, followed by reasoning over structured representations~\cite{tito2023hierarchical,zhang2024cream,luo2024layoutllm,fujitake2024layoutllm,li2024monkey,duan2025docopilot,nacson2025docvlm,xie2024wukong}. While effective on clean and well-formatted documents, these pipelines inevitably suffer from cascading OCR and layout errors and generalize poorly to noisy or out-of-domain scenarios; 2) \textbf{OCR-free Pure-Vision DocVQA} \quad Recent OCR-free methods leverage large vision--language models to reason directly over document images, preserving rich visual and spatial cues. However, scaling to long documents remains challenging. Existing approaches include: (i) \emph{end-to-end} models that process all pages jointly~\cite{hu2025mplug2,zhu2025internvl3,yan2026docseeker}, which scale poorly with document length, as computational cost and memory consumption grow rapidly with the number of pages; (ii) \emph{retrieval-based} methods that select top-$k$ pages before generation~\cite{cho2024m3docrag,yu2024visrag,chen2024sv,tanaka2025vdocrag,wang2025vrag,wu2025molorag,shi2025urag}, improving efficiency but remaining sensitive to retrieval errors and fixed hyperparameters; and (iii) \emph{agent-based} systems that iteratively explore documents~\cite{xu2025cogdoc,yang2025docagent,yue2025synergistic}, which introduce interaction at the cost of increased complexity. In contrast, our method formulates DocVQA as a sequential evidence aggregation process, enabling a single OCR-free agent to actively and efficiently aggregate visual evidence over long documents.

\begin{figure*}[t]
    \centering
    \includegraphics[width=1.0\textwidth]{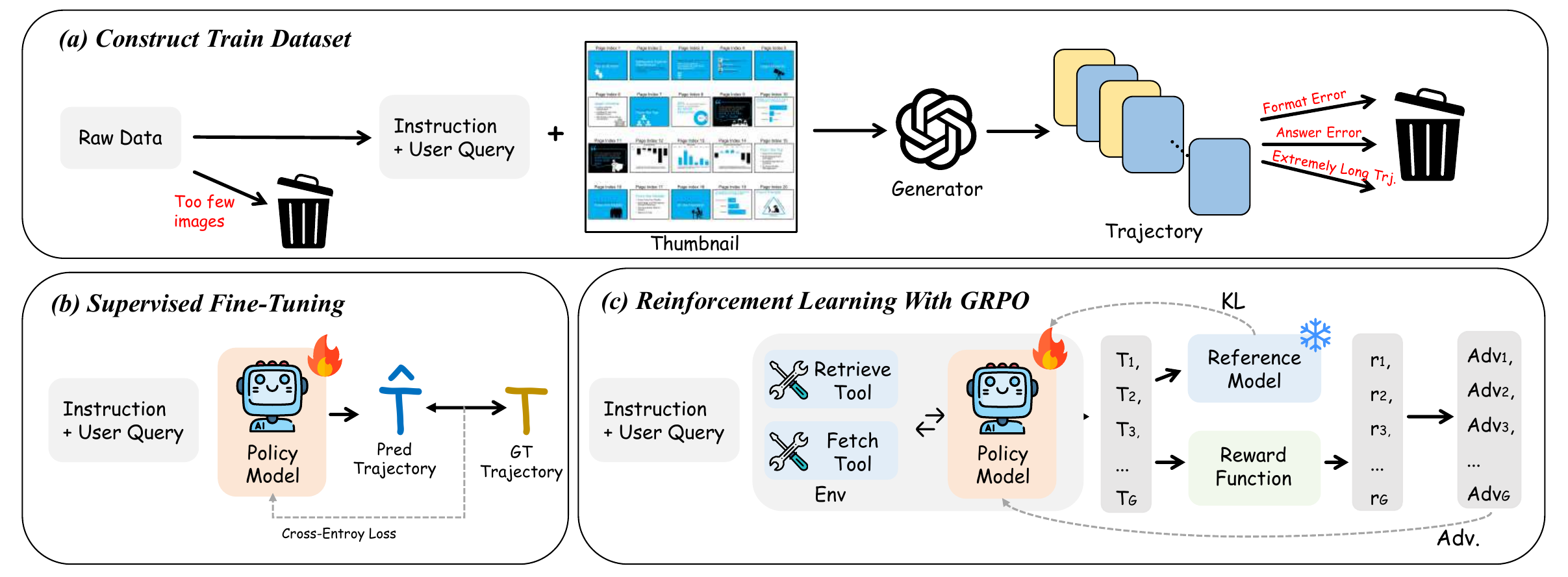}
    \vspace{-1mm}
    \caption{\textbf{Overview of the training pipeline for \textbf{Doc-$V^*$}.} \textit{(a) Training data construction.} Documents and queries are paired to generate thumbnail-guided reasoning trajectories, followed by quality filtering. \textit{(b) Supervised fine-tuning (SFT).} \textit{(c) Reinforcement learning with GRPO.} }
    \label{fig:method_overview}
\end{figure*}

\section{Method}
\label{sec:method}

\subsection{Formulation and Cognitive Motivation}

Faced with lengthy, unfamiliar documents, human experts exhibit pronounced \textbf{\emph{goal-directedness}} and \textbf{\emph{proactivity}} rather than reading cover-to-cover: they navigate using structural cues and keyword-like searches, and iteratively update their strategy as evidence is found. This behavior is consistent with \textbf{\emph{Active Vision}}~\cite{aloimonos1988active}, which views perception as goal-directed sampling to reduce uncertainty, and \textbf{\emph{Resource-Rational Cognition}}~\cite{lieder2020resource}, which trades off information gain against processing costs. Motivated by these principles, we propose \textbf{Doc-$V^*$}, formulating \emph{Multi-page Document VQA} as a \emph{Sequential Decision Process}: given a document $\mathcal{D}=\{p_1,\ldots,p_N\}$ and a question $Q$, an \textbf{OCR-free} MLLM-based agent $\pi_\theta$ interacts with the document environment for up to $T$ steps. At step $t$, the agent receives its observation $O_t$, performs reasoning, and selects an action $a_t\in\mathcal{A}$; the environment then returns feedback $E_{t+1}$, which is incorporated into the next observation $O_{t+1}$. This closed-loop formulation enables \textbf{selective evidence acquisition} and the \textbf{integration of scattered visual cues into a coherent reasoning chain}.

\subsection{Environment Design}

\noindent\textbf{Document Visual Representation} \quad Our agent is built upon the Qwen-2.5-VL~\cite{bai2025qwen25} architecture, which comprises a visual encoder $\mathcal{V}$ (adopting ViT~\cite{dosovitskiy2020image} architecture), a multi-layer perceptron projection module $\mathcal{M}$, and a large language model backbone $\mathcal{L}$. We pre-compute and cache the visual tokens $\mathbf{v}_i = \mathcal{M}(\mathcal{V}(p_i)) \in \mathbb{R}^{L_i \times d}$ for all pages \(\{p_i\}_{i=1}^N\) within \(D\) at their \textbf{native high resolution} (capped at $1024 \times 768$), where $L_i$ is the token count and $d$ is the hidden dimension. Crucially, these visual tokens are not fed to the agent all at once but are dynamically requested by the agent based on its decisions.

\noindent\textbf{Initial Observation} \quad Before interaction begins, we design a \textbf{\emph{Global Thumbnail Overview}} \(\tilde{D}\) for the document, inspired by the human behavior of first "rapidly flipping through pages" to grasp the overall structure when browsing a document. Concretely, we partition the document into groups of pages, resize each page to a thumbnail (\(256\times256\)), reorganize each group into a grid image and annotate each thumbnail with its \textbf{\emph{absolute page number}}. While body text details become indiscernible at this resolution, rich structural information remains visible like \emph{document type}, \emph{section layout}, \emph{chart distribution} and \emph{larger-font titles}. This coarse-grained global perception provides considerable navigational priors for subsequent fine-grained exploration. Formally, the initial input fed to the agent is denoted as: $O_o=\{Q,\tilde{D}\}$, where \(\tilde{D}\) possibly consisting of one or multiple grid images. 

Please refer to Appendix~\ref{app:thumbnail_overview} for the detail of the \emph{Environment Design}.

\subsection{Action Space}
\label{sec:action_space}

We define three types of atomic actions for the agent that capture common human document-reading behaviors.

\noindent\textbf{1. Retrieval Action} \quad The retrieval action is intended to approximate the "Ctrl+F search within document" behavior, but at the level of page images. To trigger this, the agent need emits a structured command: \textbf{\texttt{"<retrieval\_page>$q_t$"}}, which signifies a decision to retrieve document images using the textual query $q_t$. The query can differ from the original question \(Q\), allowing iterative refinement as evidence accumulates. The environment then calls an external multimodal retriever (e.g., ColQwen~\cite{faysse2024colpaliefficientdocumentretrieval}), ranking pages in \(\mathcal{D}\setminus\mathcal{P}_{\text{visited}}\) and returns the top-\(k\) \textbf{unvisited pages}, where $\mathcal{P}_{\text{visited}}$ is an external variable that maintains a set of visited pages to avoid redundancy.

\noindent\textbf{2. Fetch Action} \quad The fetch action requests specific pages by absolute indices via the command \textbf{\texttt{"<fetch\_page>$[i_1, \dots, i_m]$"}}. Upon receiving this, the environment parses the index list and retrieves the exact pages specified. This action facilitates several common navigation strategies: 1) direct page fetching based on visual features observed in the thumbnail view (e.g., TOC and chart positions);  2) needing to view adjacent pages before or after the current page for complete context after reading a certain page;  3) responding to page numbers explicitly mentioned in the user question (e.g., \emph{"How many baselines are there in the table on page three?"}). 

For both actions, the environment returns the \textbf{cached high-resolution visual tokens} of the requested pages. Each page's visual tokens are prefixed with a textual page number identifier (e.g., \textbf{\texttt{"Page 5:"}}) to ensure the agent can correctly associate the visual content with its specific page number. If a requested page has already been visited, the environment returns a \textbf{text reminder} instead of re-inputting the visual tokens. We denote $E_t$ as the \emph{environment feedback} at interaction step $t\ge 1$.

\noindent\textbf{3. Answer Action} \quad When the agent determines that sufficient evidence has been gathered, it terminates the interaction by executing the answer action by generating \textbf{\texttt{"<answer>\(y\)"}}, where \(y\) is the final answer string. 

\subsection{Structured Visual Reasoning}

To make the agent’s decision process explicit and auditable, we enforce a fixed \emph{think-acting} interaction protocol, a ReAct~\cite{yao2022react} reasoning style with visual feedback. At each step, the agent’s output must follow the format:
\textbf{$\texttt{"<think>}\cdots\texttt{</think>}\texttt{<action>}\cdots\texttt{</action>"}$},
where \textbf{\texttt{<action>}} instantiates exactly one action from \S\ref{sec:action_space} with the required arguments.

We further structure \textbf{\texttt{<think>}} into 3 blocks, with a slight distinction between the first turn and later turns. At turn $t{=}0$, given the initial observation, i.e., document thumbnails with question, \textbf{\texttt{<think>}} consists of: 1) \textbf{\texttt{<analysis>}}: a coarse document-level inspection from thumbnails, identifying likely question-relevant regions/pages and key visual cues; 2) \textbf{\texttt{<plan>}}: an explicit subgoal decomposition and an interaction plan, which guides subsequent actions under a limited step budget; 3) \textbf{\texttt{<summary>}}: a compact summary of the initial inspection and plan. At turns $t{>}0$, given newly returned high-resolution pages, \textbf{\texttt{<think>}} consists of: 1) \textbf{\texttt{<analysis>}}: Page-by-page content analysis of newly returned pages, evaluating each page's relevance to the user question, determining whether the evidence is sufficient to answer, and deciding on the next optimal action that can reduce uncertainty; 2) \textbf{\texttt{<relevant\_pages>}}: Explicitly outputs the list of page numbers judged to be relevant among the pages returned in the current turn. This component forces the agent to make binary relevance judgments, facilitating subsequent reward signal computation and model evaluation; 3) \textbf{\texttt{ <summary>}} An incremental information summary for the current turn, which together with historical summaries constitutes the agent's \textbf{\emph{Working Memory}}.

As interaction proceeds, image-text interleaved tokens accumulate and pages may arrive out of order, which can cause the agent to forget and drift (e.g., forgetting resolved sub-questions or repeatedly fetching a certain page). To mitigate this, we feed the agent an \textbf{augmented observation} $O_t = E_t \cup \{W_t\}, t\ge 1$, where the \textbf{\emph{Working Memory}} $W_t=\mathrm{Concat}(S_0,\ldots,S_{t-1})$ concatenates all previous \textbf{\texttt{<summary>}} within \textbf{\texttt{<think>}}. Please refer to Appendix~\ref{app:interaction_protocol} for the detail of the \emph{Agent Environment Interaction Protocol}.


\subsection{Training}

We adopt a standard two-stage training pipeline to obtain an agent that is both tool-competent and exploration-efficient under a bounded interaction budget. First, we perform supervised fine-tuning with a cross-entropy objective on distilled closed-loop interaction trajectories, where a strong teacher interacts with the real environment and we compute loss only on agent-generated tokens; we further filter trajectories by format validity, answer correctness, and evidence-page sanity, yielding \textbf{9,019} high-quality trajectories constructed from MP-DocVQA and DUDE. Second, we apply GRPO reinforcement learning using only outcome supervision: we filter \textbf{2,048} non-overlapping training examples, stratify them into easy/medium/hard buckets estimated by the SFT policy via multiple rollouts, and train the agent by sampling groups of trajectories in the same closed-loop environment and optimizing a weighted reward that combines answer correctness, evidence retrieval quality, and format validity. Full training details are provided in Appendix~\ref{app:training}.

\begin{table*}[t!]
\centering
\caption{\textbf{Comparison of different methods on five long-context and multi-page document understanding benchmarks.} The results are reported on \textbf{DUDE} (ANLS), \textbf{MPDocVQA} (ANLS), \textbf{SlideVQA} (F1), \textbf{MMLongBench-Doc} (Acc), and \textbf{LongDocURL} (Acc). ``Param.'' denotes the parameter scale (referring specifically to the \textbf{Generator} for RAG methods). ``Backbone'' indicates the underlying LLM or LVLM used. ``Paradigm'' categorizes methods into End-to-End (\textbf{E2E}), Retrieval-Augmented Generation (\textbf{RAG}), or \textbf{Agent}. The best and second-best results \textbf{among Open Source methods} are highlighted in \textbf{bold} and \underline{underlined}, respectively. Scores marked with an asterisk (${}^*$) indicate that the method's backbone was supervised fine-tuned on that specific benchmark's training set. \textcolor{red}{Red subscripts} in parentheses indicate the absolute performance gain over the baseline (Qwen2.5-VL).}
\label{tab:main_results}
\fontsize{9}{11}\selectfont
\setlength\arrayrulewidth{0.6pt}

\newcommand{\venue}[1]{~\textcolor{gray}{\scriptsize\textit{(#1)}}}

\newcommand{\ts}[1]{#1\rlap{\textsuperscript{*}}}

\newcommand{\gain}[1]{\rlap{$_{\textcolor{red}{(+#1)}}$}}

\resizebox{1\textwidth}{!}{
\begin{tabular}{l c c c c c c c c}
\toprule
\makecell[l]{\textbf{Method}} & 
\makecell{\textbf{Backbone}} & 
\makecell{\textbf{Param}} & 
\makecell{\textbf{Paradigm}} & 
\makecell{\textbf{DUDE} \\ \small (ANLS)} & 
\makecell{\textbf{MPDocVQA} \\ \small (ANLS)} & 
\makecell{\textbf{SlideVQA} \\ \small (F1)} & 
\makecell{\textbf{MMLong.} \\ \small (Acc)} & 
\makecell{\textbf{LongDoc.} \\ \small (Acc)} \\
\midrule
\multicolumn{9}{l}{\textcolor{gray}{\textit{Closed Source}}} \\
Gemini-1.5-Pro              & - & - & E2E & 46.0 & - & - & 28.2 & 50.9 \\
GPT-4o mini                 & - & - & E2E & 46.5 & - & 60.7 & 28.6 & - \\
GPT-4o                      & - & - & E2E & 54.1 & 67.4 & 65.8 & 42.8 & 64.5 \\
GPT-4.1                     & - & - & E2E & 50.2 & - & 74.7 & 45.6 & - \\
Claude-3.7-Sonnet           & - & - & E2E & 58.1 & - & 76.3 & 33.9 & - \\
\midrule
\multicolumn{9}{l}{\textcolor{gray}{\textit{Open Source}}} \\
HiVT5 \venue{PR}            & DiT / T5           & 0.3B & E2E & 23.1 & \ts{62.0} & - & - & - \\
CREAM \venue{ACM MM'24}     & Pix2Struct / LLaMa2& 7B   & RAG & \ts{52.5} & \ts{74.3} & - & - & - \\
mPLUG-DocOwl2 \venue{ACL'25}& ViT / LLaMa        & 8B   & E2E & \ts{46.8} & \ts{69.4} & 27.8 & 13.4 & 5.3 \\
M3DocRAG \venue{arXiv'24}   & Qwen2-VL           & 7B   & RAG & 39.5 & 84.4 & 55.7 & 21.0 & 35.1 \\
VisRAG \venue{ICLR'25}      & MiniCPM-V 2.6      & 8B   & RAG & 43.1 & - & 52.4 & 18.8 & 41.9 \\
SV-RAG \venue{ICLR'25}      & InternVL2          & 4B   & RAG & 45.0 & 71.0 & \ts{34.3} & 23.0 & - \\
VDocRAG \venue{CVPR'25}     & Phi3-Vision        & 4B   & RAG & \ts{44.0} & 62.6 & 42.0 & 18.4 & 39.8 \\
Docopilot \venue{CVPR'25}   & InternVL2          & 8B   & E2E & \ts{40.7} & \ts{81.3} & 43.1 & 28.8 & - \\
DocVLM \venue{CVPR'25}      & Qwen2-VL           & 7B   & E2E & 47.4 & 84.5 & - & - & - \\
InternVL3 \venue{arXiv'25}  & InternViT / Qwen2.5& 8B   & E2E & 47.4 & 80.8 & 64.4 & 24.1 & 38.7 \\
VRAG-RL \venue{NeurIPS'25}  & Qwen2.5-VL         & 7B   & Agent & - & - & - & 26.6 & 44.9 \\
MoLoRAG \venue{EMNLP'25}    & Qwen2.5-VL         & 7B   & RAG & - & - & - & \underline{41.0} & 51.9 \\
CogDoc \venue{arXiv'25}     & Qwen2.5-VL         & 7B   & Agent & \ts{46.2} & 75.0 & \ts{67.9} & 33.0 & - \\
URaG \venue{AAAI'26}        & Qwen2.5-VL         & 7B   & RAG & \ts{57.6} & \ts{\textbf{88.2}} & - & 33.8 & 52.2 \\
\midrule
\rowcolor{LLightGray} \multicolumn{9}{c}{\textit{Ours}} \\
\midrule
Qwen2.5-VL (Baseline)       & Qwen2.5-VL & 7B & E2E & 51.9 & 75.2 & 55.2 & 28.0 & 32.9 \\
Qwen2.5-VL (RAG Top-5)      & Qwen2.5-VL & 7B & RAG & 52.2\gain{0.3} & 77.4\gain{2.2} & 62.9\gain{7.7} & 36.1\gain{8.1} & 37.8\gain{4.9} \\
\textbf{Doc-$V^*$} (SFT)                & Qwen2.5-VL & 7B & Agent & \underline{58.1}\gain{6.2} & 81.3\gain{6.1} & \underline{73.8}\gain{18.6} & 39.8\gain{11.8} & \underline{53.0}\gain{20.1} \\
\textbf{Doc-$V^*$} (GRPO)               & Qwen2.5-VL & 7B & Agent & \textbf{64.5}\gain{12.6} & \underline{86.2}\gain{11.0} & \textbf{77.2}\gain{22.0} & \textbf{42.1}\gain{14.1} & \textbf{56.3}\gain{23.4} \\
\bottomrule
\end{tabular}
}
\end{table*}

\section{Experiments}
\label{sec:experiments}

\subsection{Experimental Setup}
\label{subsec:setup}

\noindent\textbf{Datasets} \quad Our raw training data is sourced from \textbf{MP-DocVQA}~\cite{tito2023hierarchical} and \textbf{DUDE}~\cite{van2023document}. Evaluation is conducted under two settings. (1) \textit{In-Domain} evaluation is performed on the test splits of MP-DocVQA and DUDE. (2) \textit{Out-of-Domain (OOD)} evaluation is carried out on three challenging benchmarks: \textbf{SlideVQA}~\cite{tanaka2023slidevqa}, \textbf{LongDocURL}~\cite{deng2025longdocurl}, and \textbf{MMLongBench-Doc}~\cite{ma2024mmlongbench}. These benchmarks cover diverse document types and reasoning challenges, enabling a comprehensive evaluation of generalization beyond the training domain. Detailed statistics and dataset characteristics are provided in Appendix~\ref{appendix:dataset}.

\noindent\textbf{Evaluation Metrics} \quad All methods are evaluated using the \emph{official metrics and evaluation protocols} of each benchmark. Specifically, we report ANLS for DUDE and MPDocVQA, F1 score for SlideVQA, and Accuracy for MMLongBench-Doc and LongDocURL.

\noindent\textbf{Agent and Environment Setup} \quad Our agent is initialized from \textbf{Qwen-2.5-VL-7B-Instruct}~\cite{bai2025qwen25}. 
For the \texttt{retrieval\_page}, we employ \textbf{ColQwen}~\cite{faysse2025colpaliefficientdocumentretrieval} as the external retriever. 
Retrieval budget is dynamically set to $k=\min(\lceil N/10\rceil,4)$ to balance information coverage and context efficiency, and the maximum interaction horizon is fixed to $T=8$ steps during both training and inference. 
The optimization objective incorporates a composite reward function balancing answer correctness ($\omega_{\text{ans}} = 0.6$), evidence recall ($\omega_{\text{evi}} = 0.3$), and structural validity ($\omega_{\text{struct}} = 0.1$). 
Specific training hyperparameters and further implementation details are provided in Appendix~\ref{sec:appendix_implementation}.


\subsection{Main Results}
\label{subsec:main_results}

We compare \textbf{Doc-$V^*$} with a broad suite of baselines spanning three paradigms: (i) \textbf{End-to-End (E2E)} models including HiVT5~\cite{tito2023hierarchical}, mPLUG-DocOwl2~\cite{hu2025mplug2}, Docopilot~\cite{duan2025docopilot}, DocVLM~\cite{nacson2025docvlm}, and InternVL3~\cite{zhu2025internvl3}; (ii) \textbf{Retrieval-Augmented Generation (RAG)} methods including CREAM~\cite{zhang2024cream}, M3DocRAG~\cite{cho2024m3docrag}, VisRAG~\cite{yu2024visrag}, SV-RAG~\cite{chen2024sv}, VDocRAG~\cite{tanaka2025vdocrag}, MoLoRAG~\cite{wu2025molorag}, and URaG~\cite{shi2025urag}; and (iii) \textbf{Agent-based} approaches including VRAG-RL~\cite{wang2025vrag} and CogDoc~\cite{xu2025cogdoc}. We additionally report closed-source systems (Gemini-1.5-Pro~\cite{team2024gemini}, GPT-4o mini, GPT-4o~\cite{hurst2024gpt}, GPT-4.1, and Claude-3.7-Sonnet) as reference points, and include Qwen2.5-VL~\cite{bai2025qwen25} along with its RAG Top-5 variant as direct backbone baselines. Detailed descriptions of these baseline methods are provided in Appendix~\ref{sec:appendix_baseline}. As shown in Table~\ref{tab:main_results}, our GRPO-enhanced model achieves the best overall performance among open-source methods on four of five benchmarks, while remaining competitive on the remaining benchmark.

On the \textbf\emph{{In-domain}} benchmarks (DUDE and MPDocVQA), \textbf{Doc-$V^*$} achieves strong accuracy. On DUDE, it reaches \textbf{64.5} ANLS, \textbf{outperforming all open-source baselines} and also \textbf{surpassing some closed-source models reported}, including GPT-4o (54.1) and Claude-3.7-Sonnet (58.1). On MPDocVQA, our method attains \textbf{86.2} ANLS, remaining highly competitive with URaG (88.2).

On the \textbf\emph{{Out-of-Domain}} benchmarks, \textbf{Doc-$V^*$} shows clear generalization advantages. On SlideVQA, our model achieves 77.2 F1, outperforming SlideVQA-trained baselines CogDoc (67.9). It also sets new open-source highs on long-context benchmarks, scoring 42.1 accuracy on MMLongBench-Doc and 56.3 accuracy on LongDocURL. These results indicate that \textbf{Doc-$V^*$} maintains robust long-context evidence localization and aggregation ability when transferring to diverse document domains and substantially longer inputs.

To isolate the effect of the agentic policy and GRPO training, we compare \textbf{Doc-$V^*$} against Qwen2.5-VL and Qwen2.5-VL (RAG Top-5) under the same 7B scale. Static retrieval is beneficial—Qwen2.5-VL (RAG Top-5) improves over the vanilla backbone, e.g., 28.0 $\rightarrow$ 36.1 on MMLongBench-Doc and 32.9 $\rightarrow$ 37.8 on LongDocURL. Nevertheless, our proposed method yields substantially larger gains at the same parameter scale, improving over RAG Top-5 by +12.3 on DUDE (52.2 $\rightarrow$ 64.5) and +18.5 on LongDocURL (37.8 $\rightarrow$ 56.3). These results demonstrate that optimizing a multi-step evidence-seeking policy via GRPO offers superior robustness compared to fixed top-$k$ retrieval, allowing small open-source models to rival powerful closed-source models in complex document understanding.


\begin{figure*}[t]
  \centering
  \includegraphics[width=0.95\linewidth]{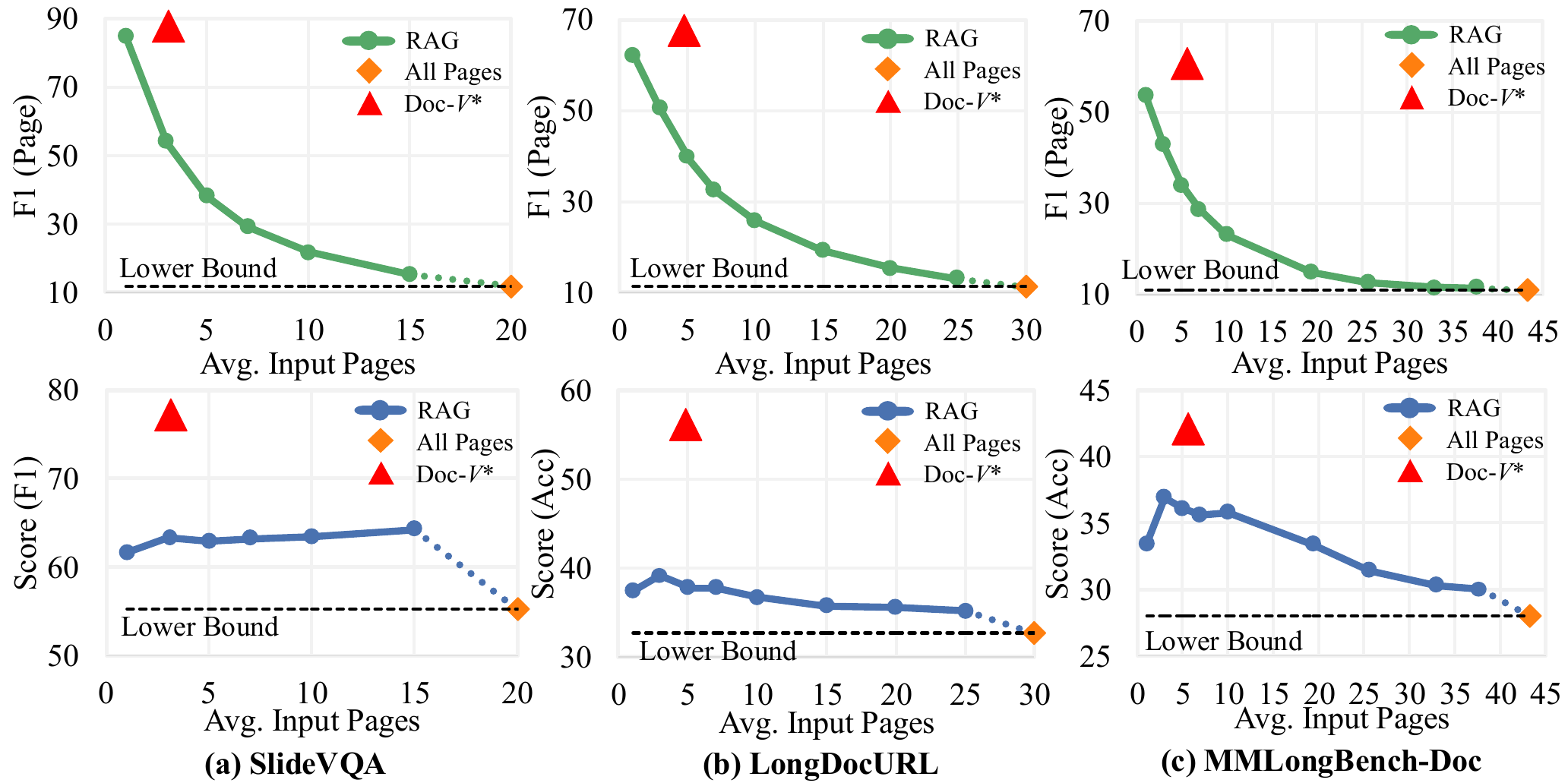}
  \caption{Efficiency--effectiveness trade-off across \textbf{SlideVQA}, \textbf{LongDocURL}, and \textbf{MMLongBench-Doc}.The top row reports Page-F1, measuring the quality of page selection under different input budgets, while the bottom row shows downstream task performance. \textbf{For \textbf{Doc-$V^*$}}, Page-F1 is computed based on the pages that the model explicitly predicts as relevant, i.e., the model outputs a set of \texttt{relevant\_pages}, which are then compared against the ground-truth evidence pages to compute F1.}
  \label{fig:efficiency_tradeoff}
\end{figure*}

\subsection{Analysis of Page-Level Retrieval}
\label{subsec:retrieval_analysis}

Figure~\ref{fig:efficiency_tradeoff} analyzes the trade-off between the average number of input pages and both page-level evidence quality and downstream task performance. This analysis directly probes how different methods handle evidence under constrained budgets.

For \textbf{multimodal RAG}, increasing the number of retrieved pages exhibits a characteristic non-monotonic trend: performance initially improves as more relevant pages are included, but degrades once additional pages introduce noise. This behavior highlights two structural limitations of RAG-style pipelines. First, performance is highly sensitive to the choice of \textbf{Top-$K$}. Second, evidence selection and reasoning are loosely coupled—the generator must attend over a fixed, noisy context without explicit mechanisms for evidence validation or revision.

In contrast, \textbf{Doc-$V^*$} frames long-document understanding as a \textbf{progressive evidence aggregation process}. Instead of consuming all pages at once, the agent incrementally explores the document, extracts candidate evidence, and explicitly decides which pages are relevant at each step. This difference is reflected in the \textbf{Page-F1 metric}, which measures the alignment between the pages ultimately selected by the model and the ground-truth evidence pages.

Under comparable average input budgets, \textbf{Doc-$V^*$} consistently achieves substantially higher Page-F1 than RAG across SlideVQA, LongDocURL, and MMLongBench-Doc. Importantly, this improvement does not arise from retrieving more pages, but from \textbf{selectively consolidating evidence across multiple interaction steps}. Early observations guide hypothesis formation, while later page accesses serve to verify, refine, or reject these hypotheses.

These results suggest that \textbf{long-document understanding is not limited by insufficient context, but by the model’s ability to organize and integrate evidence}. Revisiting the behavior of multimodal RAG, increasing the number of input pages primarily amplifies irrelevant or weakly related signals, while lacking explicit mechanisms for evidence consolidation. As a result, evidence becomes diluted rather than reinforced, leading to degraded reasoning performance.

\subsection{Robustness Analysis}
\label{sec:robustness}

In this section, we analyze the robustness of our framework regarding the number of reasoning steps and the efficiency trade-off compared to traditional retrieval methods. More analysis see Appendix~\ref{appendix:robustness}

\begin{table}[t]
\centering
\caption{\textbf{Comparison of different retrievers on MMLongBench-Doc.}}
\label{tab:retriever_comparison}
\footnotesize
\setlength{\tabcolsep}{4pt}
\renewcommand{\arraystretch}{1.1}
\resizebox{\columnwidth}{!}{
\begin{tabular}{c c c c c c c c}
\toprule
\textbf{Retriever} & \textbf{Model} & \textbf{Avg. Pages} & \textbf{Page-F1} & \textbf{Overall} & \textbf{SIN} & \textbf{MUL} & \textbf{UNA} \\
\midrule
\multirow{2}{*}{ColQwen} 
& Qwen2.5-VL    & 6.0 & 30.9 & 35.5 & 37.0 & 13.4 & \textbf{70.4} \\
& \textbf{Doc-$V^*$}        & 5.6 & \textbf{49.7} & \textbf{42.1} & \textbf{54.6} & \textbf{23.5} & 45.7 \\
\midrule
\multirow{2}{*}{BGE-Large} 
& Qwen2.5-VL    & 9.0 & 17.6 & 33.0 & 31.2 & 9.8  & \textbf{77.1} \\
& \textbf{Doc-$V^*$}        & 8.4 & \textbf{34.0} & \textbf{36.3} & \textbf{45.7} & \textbf{18.5} & 45.7 \\
\midrule
\multirow{2}{*}{BM25} 
& Qwen2.5-VL    & 10.0 & 20.5 & 32.9 & 32.7 & 11.3 & \textbf{71.3} \\
& \textbf{Doc-$V^*$}        & 9.2 & \textbf{36.8} & \textbf{37.5} & \textbf{48.4} & \textbf{20.4} & 43.0 \\
\bottomrule
\end{tabular}
}
\end{table}

\begin{figure}[t]
    \centering
    \includegraphics[width=\linewidth]{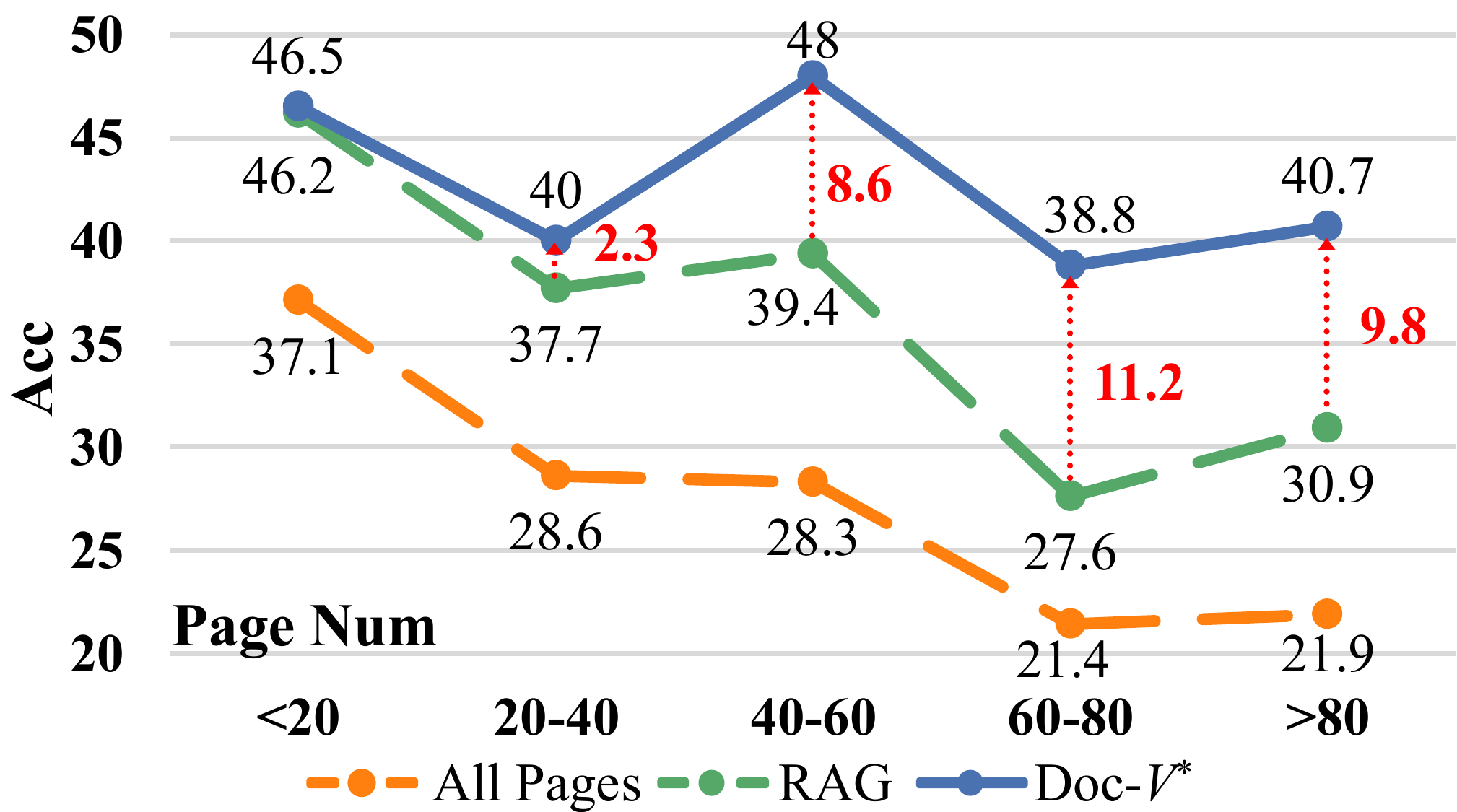} 
    \caption{Accuracy vs. document length under different methods (RAG uses top-k = 5 retrieval).}
    \label{fig:doc_length}
\end{figure}

\noindent\textbf{Impact of Document Length} \quad Figure~\ref{fig:doc_length} shows performance across different document length ranges.  
Both \textit{All Pages} and \textit{RAG} exhibit a clear performance degradation as document length increases, whereas \textbf{Doc-$V^*$} maintains consistently strong results across all ranges. Both \textit{All Pages} and \textit{RAG} suffer from substantial performance degradation as document length increases, while \textbf{Doc-$V^*$} remains consistently strong. 
In the longest-document regime ($>80$ pages), \textbf{Doc-$V^*$} outperforms RAG by \textbf{31.7\%} (40.7 vs.\ 30.9) and exceeds the \textit{All Pages} setting by a large margin of \textbf{85.8\%} (40.7 vs.\ 21.9), demonstrating its \textbf{effectiveness} and \textbf{robustness} for long-document understanding.

\begin{figure}[t]
    \centering
    \includegraphics[width=0.95\linewidth]{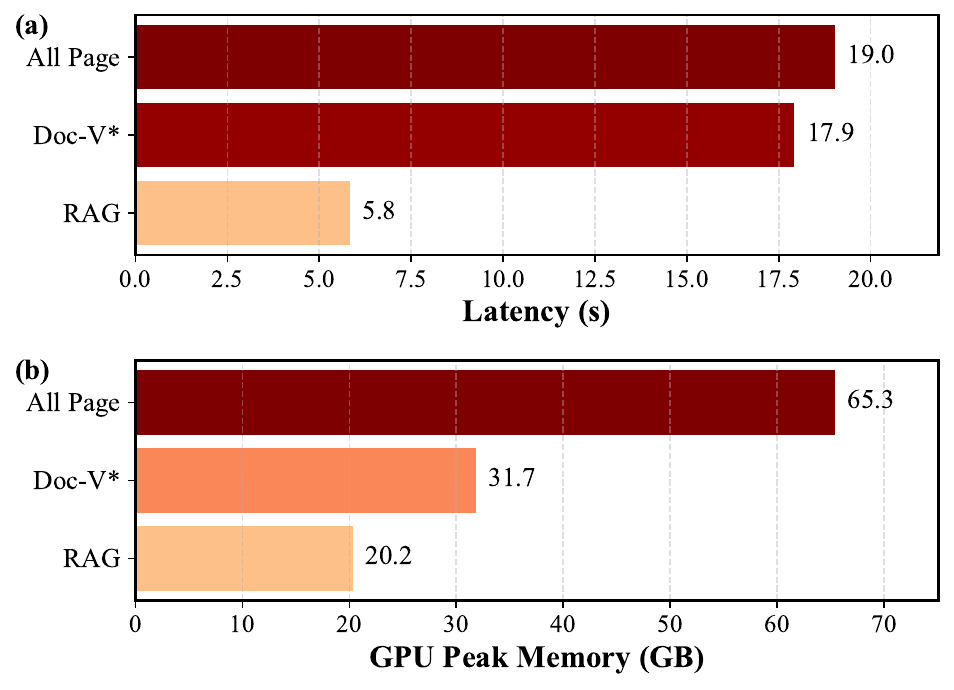} 
    \caption{\textbf{(a)}: Comparison of average inference latency per sample across different methods. \textbf{(b)}: Comparison of average peak GPU memory consumption per sample under different methods.}
    \label{fig:efficiency_cost}
\end{figure}

\noindent\textbf{Efficiency and Cost} \quad To evaluate the efficiency and computational cost of different document processing strategies, a subset of samples with long documents is randomly selected for analysis. 
These samples are characterized by a large number of pages, with an average document length of \textbf{107.3} pages, which provides a representative setting for assessing scalability under realistic long-document scenarios.
Figure~\ref{fig:efficiency_cost} presents a comparative analysis of inference latency and GPU memory consumption across different methods.
The results indicate that processing the entire document at once leads to substantially higher inference latency and GPU memory consumption, as all pages must be loaded and processed simultaneously.
By contrast, the standard RAG baseline significantly reduces both latency and memory footprint by restricting computation to a small subset of retrieved pages.
\textbf{Doc-$V^*$} occupies a middle ground between these two extremes: while incurring higher cost than RAG due to iterative page access and multi-step reasoning, it avoids the prohibitive overhead of full-document processing and achieves a more favorable balance between efficiency and document coverage.

\noindent\textbf{Impact of Different Retrievers} \quad Table~\ref{tab:retriever_comparison} shows that \textbf{Doc-$V^*$} maintains strong overall performance across retrievers with substantially different capabilities.
Even when coupled with weak text-based retrievers (BM25~\cite{robertson2009probabilistic}, BGE-Large~\cite{bge_embedding}), which suffer from low Page-F1 and increased noise due to OCR and layout loss, Doc-$V^*$ incurs only moderate performance degradation, indicating limited dependence on high-quality retrieval. Unlike conventional RAG pipelines where downstream performance is tightly coupled with retrieval recall, this robustness stems from \textbf{Doc-$V^*$}’s active compensation mechanism: when initial retrieval misses critical evidence, the model detects contextual insufficiency and proactively recovers missing pages via browsing actions (e.g., \texttt{fetch\_page}), effectively acting as an intelligent correction layer rather than a passive consumer.




\subsection{Ablation Study}

To validate the design choices of the proposed agentic framework, we conduct ablation experiments on \textbf{MMLongBench-Doc}, focusing on both the cognitive modules that govern the agent’s reasoning process and the navigation actions that support evidence acquisition.

\noindent\textbf{Importance of Multi-granularity Page Understanding} \quad Removing either the global thumbnail overview or the page-by-page analysis module causes significant performance drops of 4.9 and 4.7 accuracy points, respectively (Table~\ref{tab:module_ablation}), indicating that effective long-document reasoning relies on multi-granularity page understanding. The global overview provides structural cues for efficient navigation, while fine-grained analysis enables precise evidence extraction; using only one level of perception leads to either inefficient exploration or insufficient evidence recovery.

\noindent\textbf{Complementary Roles of Retrieval and Fetch Actions} \quad
We analyze the agent’s navigation behavior using both page-level metrics (Table~\ref{tab:page_retrieval_analysis}) and action ablation on MMLongBench-Doc (Table~\ref{tab:action_ablation}). 
The \texttt{retrieval\_page} action achieves higher recall but lower precision, serving as a coarse semantic filter, while \texttt{fetch\_page} provides higher precision for fine-grained evidence grounding. Ablation results further confirm their complementarity: removing retrieval leads to inefficient exploration (more pages), while removing fetch degrades accuracy.
Combining both yields the best accuracy–efficiency trade-off, forming a coarse-to-fine evidence aggregation strategy.

\begin{table}[t]
\centering
\caption{Ablation study on the cognitive modules of the \textbf{Doc-$V^*$} agent. \textbf{T}: Global Thumbnail Overview; \textbf{A}: Page-by-page content analyis; \textbf{M}: Memory.}
\label{tab:module_ablation}
\footnotesize
\setlength{\tabcolsep}{4pt}
\renewcommand{\arraystretch}{1.1}
\begin{tabular}{c c c | c c c }
\toprule
\multicolumn{3}{c|}{\textbf{Cognitive Modules}} & \textbf{MMLong.}  & \textbf{LongDoc.} & \textbf{SlideVQA}\\
\textbf{T}  & \textbf{A} & \textbf{M} & (Acc) & (Acc) & (F1)\\
\midrule
\rowcolor{gray!10}
\checkmark & \checkmark & \checkmark & \textbf{39.8} & \textbf{53.0} & \textbf{73.8} \\
\midrule
\xmark & \cmark & \cmark & $34.9_{\textcolor{darkgreen}{\textbf{(\text{-}4.9)}}}$ & $46.3_{\textcolor{darkgreen}{\textbf{(\text{-}6.7)}}}$ & $68.3_{\textcolor{darkgreen}{\textbf{(\text{-}5.5)}}}$\\
\cmark & \xmark & \cmark & $35.9_{\textcolor{darkgreen}{\textbf{(\text{-}4.7)}}}$ & $49.5_{\textcolor{darkgreen}{\textbf{(\text{-}3.5)}}}$ &
$71.8_{\textcolor{darkgreen}{\textbf{(\text{-}2.0)}}}$\\
\cmark & \cmark & \xmark  & $36.4_{\textcolor{darkgreen}{\textbf{(\text{-}3.4)}}}$ & $47.1_{\textcolor{darkgreen}{\textbf{(\text{-}5.9)}}}$ &
$69.8_{\textcolor{darkgreen}{\textbf{(\text{-}4.0)}}}$\\
\bottomrule
\end{tabular}
\end{table}

\begin{table}[t]
\centering
\caption{
Action ablation study on MMLongBench-Doc.
Removing either retrieval or fetch leads to clear performance degradation.
}
\begin{tabular}{lcc}
\toprule
\textbf{Setting}  & \textbf{Acc} $\uparrow$ & \textbf{Avg. Pages} $\downarrow$ \\
\midrule
Doc-$V^*$  & \textbf{39.8} & 6.4 \\
w/o Retrieval  & 34.9 & 14.2 \\
w/o Fetch  & 35.2 & \textbf{5.9} \\
\bottomrule
\end{tabular}
\label{tab:action_ablation}
\end{table}

\begin{table}[t]
\centering
\caption{\textbf{Page-level analysis of agent tool usage and retrieval quality across three benchmarks.} \textbf{RP} denotes pages retrieved by the \texttt{retrieval\_page}, while \textbf{FP} denotes pages obtained via the \texttt{fetch\_page}. \textbf{Ratio} indicates the proportion of samples in which the corresponding tool is invoked. \textbf{Recall}, \textbf{Precision}, and \textbf{F1} are computed at the page level.
}
\small
\setlength{\tabcolsep}{6pt}
\renewcommand{\arraystretch}{1.15}
\begin{tabular}{lcccccc}

\toprule
\multirow{2}{*}{\textbf{Metric}} 
 & \multicolumn{2}{c}{\textbf{SlideVQA}} 
 & \multicolumn{2}{c}{\textbf{LongDoc.}} 
 & \multicolumn{2}{c}{\textbf{MMLong.}} \\
\cmidrule(lr){2-3} \cmidrule(lr){4-5} \cmidrule(lr){6-7}
 & RP & FP & RP & FP & RP & FP \\
\midrule
Ratio      & 97.6 & 4.1  & 99.8 & 3.6  & 94.0 & 14.7 \\
Recall     & 95.7 & 70.9 & 83.4 & 37.3 & 75.4 & 55.9 \\
Precision  & 39.0 & 81.2 & 32.7 & 36.6 & 33.1 & 49.9 \\
F1         & 54.1 & 72.9 & 44.4 & 31.9 & 42.1 & 49.6 \\
\bottomrule
\end{tabular}
\label{tab:page_retrieval_analysis}
\end{table}

\section*{Conclusion}
This paper introduces \textbf{Doc-V$^\ast$}, an OCR-free agentic framework for multi-page document VQA via active evidence aggregation. Experiments on five benchmarks show gains over strong open-source baselines and competitive results against proprietary models, particularly on long and OOD documents. These findings position selective evidence aggregation as a robust alternative to fixed-context and retrieval-augmented methods.

\section*{Limitations}
This work is subject to several limitations. First, all experiments are conducted with a single backbone (Qwen2.5-VL), and the effectiveness of the proposed agentic framework across different vision–language backbones is not systematically evaluated. Although the method is conceptually backbone-agnostic, architectural differences may affect evidence aggregation and tool usage behaviors. Second, Doc-$V^*$ is evaluated only in the single-document setting; its performance on multi-document scenarios, where evidence must be aggregated across multiple heterogeneous documents, remains unexplored and requires further study.

\section*{Ethical Considerations}
Most datasets used in this work are publicly available benchmarks for document visual question answering and are utilized in accordance with their respective licenses. The proposed framework does not introduce new data collection or annotation processes involving human subjects. Similar to existing vision–language models, Doc-$V^*$ may produce incorrect or incomplete answers due to hallucination or imperfect evidence aggregation, particularly on complex or ambiguous documents. As with prior work, its outputs are intended to support document understanding and analysis, rather than to serve as authoritative or final interpretations.

\section*{Acknowledgments}
This work is supported by the NSFC (62225603).

\bibliography{custom}

\appendix

\section{Detail of Environment Design}
\label{app:thumbnail_overview}

This subsection provides the detailed construction of the \textbf{\emph{Global Thumbnail Overview}} \(\tilde{D}\) referenced in our \emph{Environment Initialization}. Given a document with \(N\) pages \(D=\{I_1,\ldots,I_N\}\), we build \(\tilde{D}\) as a small set of tiled overview images that together cover all pages while maintaining a very low initial visual budget compared to all image with high-resolution. We set \(G=36\) to be the maximum number of pages allowed per overview image. We first partition the page indices into consecutive groups in sequential order
\[
\mathcal{G}_k=\{(k-1)G+1,\ldots,\min(kG,N)\},
\]
where $ k=1,\ldots,K$, and the number of overview images $K$ is
\[
K=\left\lceil\frac{N}{G}\right\rceil,\quad n_k = |\mathcal{G}_k| \le G.
\]
Each page \(I_i\) is resized to a fixed thumbnail \(T_i\in\mathbb{R}^{256\times256}\) (aspect-ratio handling follows standard padding/letterboxing so that all thumbnails share identical canvas size). For each group \(\mathcal{G}_k\), we pack its \(n_k\) thumbnails into a single composite image \(\tilde{I}^{(k)}\) using an adaptive near-square grid. Concretely, we choose grid dimensions \((R_k,C_k)\) such that \(R_k C_k \ge n_k\) and the grid is as close to square as possible; in practice, we set
\[
R_k = \left\lceil \sqrt{n_k} \right\rceil,\quad C_k = \left\lceil \frac{n_k}{R_k} \right\rceil,
\]
which guarantees \(R_k C_k \ge n_k\) and yields a compact layout. If \(R_k C_k > n_k\), the remaining cells are left empty (blank padding) to preserve a regular grid geometry.

To ensure unambiguous visual indexing, each grid cell includes a thin blank header band of height \(h\) pixels above the thumbnail region; we render the absolute page index \(i\) (for the corresponding thumbnail \(T_i\)) inside this header band. Thus, a cell is a \((h+256)\times256\) block consisting of a header strip for the index and a \(256\times256\) thumbnail area below it. The resulting overview image \(\tilde{I}^{(k)}\) is obtained by tiling these blocks into an \(R_k\times C_k\) array, with empty cells rendered as blank blocks.

This construction yields the global overview set
\[
\tilde{D}=\{\tilde{I}^{(1)},\ldots,\tilde{I}^{(K)}\},\quad K=\left\lceil\frac{N}{G}\right\rceil,
\]
which is then used in the initial observation \(O_1=\{Q,\tilde{D}\}\) as described in the main paper.

For intuition, consider several document lengths. When \(N=40\), we obtain \(K=\lceil 40/36\rceil=2\) overview images: the first group has \(n_1=36\) pages and forms a \(6\times 6\) grid, while the second group has \(n_2=4\) pages and forms a \(2\times 2\) grid. When \(N=50\), we again have \(K=2\): the first overview remains \(6\times6\) (36 pages), and the second overview contains \(n_2=14\) pages, which under the near-square rule becomes a \(4\times4\) grid with two empty cells. In the appendix~\ref{sec:case_study}, we visualizes these overviews images, it illustrates that these low-cost overviews provide strong initial navigational signals, especially for counting-style user questions. 

In summary, a critical advantage of the proposed \textbf{\emph{Global Thumbnail Overview}} is the substantial reduction in visual token consumption compared to full-resolution ingestion. Empirical analysis using the Qwen-2.5-VL~\cite{bai2025qwen25} vision encoder demonstrates that our method achieves a compression ratio of approximately $10\times$ to $12\times$. For instance, a 100-page document processed at a standard high resolution of $1024\times768$ typically generates over $100,000$ visual tokens. In contrast, representing the same document via our tiled overview construction (resulting in $K=3$ composite images) yields only $\approx 8,000$ visual tokens. While further downscaling of individual page thumbnails $T_i$ is theoretically possible, our chosen resolution strikes a balance between \textbf{legibility and efficiency}. Consequently, this approach functions as a strategic compromise between full-document input—which preserves global context but incurs prohibitive computational costs—and Visual Retrieval-Augmented Generation (RAG), which optimizes for cost but often fragments global coherence. By retaining a macro-level visual representation, we preserve structural and semantic continuity while leveraging external tools for fine-grained details. 

\section{Agent--Environment Interaction Protocol}
\label{app:interaction_protocol}

This section provides a complete, implementation oriented description of how the \textbf{Doc-$V^*$} agent interacts with a multi-page document environment. Our goal is to make the interaction loop explicit and reproducible: what the agent \emph{receives} at each turn, what it \emph{must output}, how the environment \emph{responds}, and how state (e.g., visited pages and working memory) is maintained. Please refer to Algorithm~\ref{alg:docvstar_loop} for the complete pseudocode.


Given a document $\mathcal{D}=\{p_1,\ldots,p_N\}$ (each $p_i$ is a page image) and a question $Q$, we cast multi-page Document VQA as a sequential decision process with a maximum budget of $T$ interaction turns. At each turn $t$:
(i) the agent receives an observation $O_t$,
(ii) it performs reasoning and emits exactly one atomic action $a_t\in\mathcal{A}$,
(iii) the environment executes the action and returns feedback $E_{t+1}$,
(iv) the feedback is incorporated into the next observation.

Crucially, the agent is \textbf{not} given the full document at high resolution upfront. Instead, the environment pre-computes and caches high-resolution visual tokens for each page and only reveals the requested pages on demand, enabling selective evidence acquisition under limited context/computation budgets.


\begin{algorithm}
\caption{\textbf{Doc-$V^*$} Agent--Environment Interaction (Inference-Time Loop)}
\label{alg:docvstar_loop}
\small
\begin{algorithmic}[1]
\Require Document pages $\mathcal{D}=\{p_1,\ldots,p_N\}$, question $Q$, turn limit $T$, retrieval top-$k$, Global Thumbnail Overview $\tilde{D}$, high-res visual tokens $\mathbf{v}_i \gets \mathcal{M}(\mathcal{V}(p_i))$
\Statex \textbf{Initialization:}
\State $\mathcal{P}_{\text{visited}} \gets \emptyset$
\Comment{tracks pages already revealed to the agent}
\State $W \gets \emptyset$
\Comment{working memory: concatenated per-turn summaries}
\State $O \gets \{Q,\tilde{D}\}$
\Comment{initial observation $O_0$}

\For{$t \gets 0$ to $T-1$}
    \State $u_t \gets \pi_\theta(O)$
    \Comment{must follow \texttt{<think>...</think><action>...</action>}}
    \State Parse $u_t$ to obtain (i) one atomic action $a_t$ and (ii) summary $S_t$
    \State $W \gets W \oplus S_t$
    \Comment{append summary to working memory}
    \If{$a_t$ is \texttt{<answer>} with string $y$}
        \State \Return $y$
        \Comment{terminate interaction}
    \ElsIf{$a_t$ is \texttt{<retrieval\_page>} with query $q_t$}
        \State $\mathcal{I} \gets \textsc{Retriever}(q_t,\ \mathcal{D}\setminus\mathcal{P}_{\text{visited}},\ k)$
        \Comment{rank \emph{unvisited} pages using an external multimodal retriever}
    \ElsIf{$a_t$ is \texttt{<fetch\_page>} with indices $[i_1,\ldots,i_m]$}
        \State $\mathcal{I} \gets [i_1,\ldots,i_m]$
        \Comment{direct request by absolute page indices}
    \Else
        \State $\mathcal{I} \gets \emptyset$
        \Comment{invalid action; environment may return a format reminder}
    \EndIf

    \Statex \textbf{Environment feedback construction:}
    \State $E \gets \emptyset$
    \ForAll{$i \in \mathcal{I}$}
        \If{$i \in \mathcal{P}_{\text{visited}}$}
            \State $E \gets E \cup \{\texttt{``Page $i$ already visited.''}\}$
            \Comment{avoid redundant visual tokens}
        \Else
            \State $E \gets E \cup \{\texttt{``Page $i$:''},\ \mathbf{v}_i\}$
            \Comment{prefix page id + cached high-res tokens}
            \State $\mathcal{P}_{\text{visited}} \gets \mathcal{P}_{\text{visited}} \cup \{i\}$
        \EndIf
    \EndFor

    \State $O \gets E \cup \{W\}$
    \Comment{augmented observation for next turn: $O_{t+1}$}
\EndFor
\State \Return \texttt{NoAnswer}
\Comment{optional fallback when turn budget is exhausted}
\end{algorithmic}
\end{algorithm}

\noindent\textbf{Cached high-resolution page tokens} \quad For each page $p_i$, the environment caches its high-resolution visual tokens
$\mathbf{v}_i=\mathcal{M}(\mathcal{V}(p_i))\in\mathbb{R}^{L_i\times d}$,
computed at the page's native resolution (capped at $1024\times768$).

\noindent\textbf{Initial Observation ($t{=}0$).} \quad Before any interaction, the environment constructs a \emph{Global Thumbnail Overview} $\tilde{D}$ by resizing pages to thumbnails (e.g., $256\times256$), arranging them into one or more grid images, and annotating each thumbnail with its \emph{absolute page number}. While fine text is typically unreadable at this scale, it preserves strong structural cues (document type, section layout, chart distribution, large-font titles). The initial observation is
\[
O_0 = \{Q, \tilde{D}\}.
\]

\noindent\textbf{Visited Page Set} \quad The environment maintains an external set $\mathcal{P}_{\text{visited}}$ to prevent redundant page inputs. If the agent requests an already visited page, the environment returns a short \emph{text reminder} rather than re-sending visual tokens.

\noindent\textbf{Working Memory} \quad To reduce forgetting and repetitive behaviors during multi-turn interaction, we maintain a \emph{Working Memory} $W_t$ formed by concatenating the agent's per-turn summaries:
\[
W_t = \mathrm{Concat}(S_0,\ldots,S_{t-1}),
\]
where $S_t$ is the content of the agent's \texttt{<summary>} block at turn $t$.

\noindent\textbf{Augmented Observation ($t{\ge}1$).} \quad At turn $t\ge 1$, the agent receives an augmented observation:
\[
O_t = E_t \cup \{W_t\},
\]
where $E_t$ is the environment feedback produced by executing the previous action.


At each turn, the agent must output \emph{exactly one} atomic action from the following set:
\begin{itemize}
    \item \textbf{Retrieval action:} \texttt{<retrieval\_page> $q_t$}.  
    This action mimics a ``Ctrl+F''-like search but over page images. The query $q_t$ may differ from the original question $Q$ and can be iteratively refined.
    \item \textbf{Fetch action:} \texttt{<fetch\_page>[$i_1,\ldots,i_m$]}.  
    This action requests pages by absolute indices (e.g., based on thumbnail cues, adjacency exploration, or explicit page references in the question).
    \item \textbf{Answer action:} \texttt{<answer> $y$}.  
    This action terminates the interaction and outputs the final answer string $y$.
\end{itemize}

To make decision-making auditable, we enforce a fixed ReAct-style output schema:
\[
\texttt{<think>...</think><action>...</action>}.
\]
At $t{=}0$, the \texttt{<think>} section should include (i) \texttt{<analysis>} based on thumbnails, (ii) \texttt{<plan>} for a turn-budgeted strategy, and (iii) \texttt{<summary>} to be appended to working memory.  
At $t{>}0$, the \texttt{<think>} section should include (i) \texttt{<analysis>} of newly returned pages, (ii) \texttt{<relevant\_pages>} listing the page numbers judged relevant among the newly returned pages, and (iii) \texttt{<summary>}.

\paragraph{Environment Response Semantics}

For retrieval or fetch actions, the environment returns the cached high-resolution visual tokens of the requested pages. Each page's tokens are preceded by a textual page identifier (e.g., \texttt{``Page 5:''}) to maintain an unambiguous mapping between content and absolute page index, especially when pages arrive out of order. For already-visited pages, the environment returns a short reminder string instead of re-injecting tokens.

\section{Detail of Training}
\label{app:training}

Existing \emph{Multi-page Document Visual Question Answering (VQA)} benchmarks usually annotate only the final supervision tuple $(D,Q,y,P_{\mathrm{gt}})$, i.e., the document, question, final answer, and (optionally) evidence pages, but they do not provide the multi-step interaction traces required by our agent. To train the behavior model described in the main text, we adopt a \textbf{two-stage recipe}: first supervised fine-tuning (SFT) on distilled closed-loop interaction trajectories, and then GRPO-based~\cite{guo2025deepseek} reinforcement learning to further optimize answer correctness and evidence discovery under a bounded interaction budget. In both stages, all environment feedback (returned page images and working memorys) is used only as conditioning context; training losses are applied only to tokens generated by the agent itself. 

\paragraph{\textbf{SFT: Closed-loop Interaction Trajectory Distillation}}
We distill interaction trajectories from a strong teacher model (GPT-4o~\cite{hurst2024gpt}) by running it in a closed-loop environment that executes real actions and returns real page images. Each teacher turn must follow our protocol: one \textbf{\texttt{<think>}} block plus exactly one \textbf{\texttt{<action>}} among \textbf{\texttt{<retrieval\_page>}}, \textbf{\texttt{<fetch\_page>}}, and \textbf{\texttt{<answer>}}. The environment executes the action and returns the corresponding visual observation (thumbnail overview at the beginning; high-resolution pages thereafter) and working memory as feedback for the next turn. This closed-loop distillation is essential because retrieval and fetching change subsequent observations, so the distilled traces reflect realistic exploration dynamics rather than offline labels.

\paragraph{\textbf{SFT: Trajectory Filtering}}
We keep only reliable trajectories for imitation: 1) \textbf{Format validity:} the full trace must be parseable; every turn contains exactly one valid action with valid arguments and required fields in \textbf{\texttt{<think>}}; 2) \textbf{Answer correctness:} we compare the teacher final answer $\hat{y}$ with the ground-truth $y$. For free-form textual answers, we compute ANLS and require $\mathrm{ANLS}(\hat{y},y)\ge \tau_{\mathrm{anls}}$ (we use $\tau_{\mathrm{anls}}=0.7$). For identifier-like answers (dates, counts, phone numbers, emails), we require exact match $\mathbb{I}[\hat{y}=y]=1$. When ANLS is low (may be due to benign formatting differences), we additionally use a judge model (GPT-4o) to verify semantic equivalence; 3) \textbf{Evidence sanity:} the teacher outputs \textbf{\texttt{<relevant\_pages>}} inside \textbf{\texttt{<think>}}. Let $P_{\mathrm{rel}}$ be the union of all pages listed in \textbf{\texttt{<relevant\_pages>}} across turns. We require $P_{\mathrm{rel}}\cap P_{\mathrm{gt}}\neq\emptyset$; if not, we keep the trajectory only if another judge model (GPT-4o) verifies that the selected pages support the answer (to mitigate incomplete evidence annotations). 

We build long-document training samples by selecting examples with more than 10 pages from MP-DocVQA~\cite{tito2023hierarchical} and DUDE~\cite{van2023document} dataset. We keep DUDE \texttt{not-answerable} cases to improve abstention when evidence is insufficient. After distillation and filtering, our SFT set contains 9{,}019 trajectories in total (5{,}969 from MP-DocVQA and 3{,}050 from DUDE).
Each distilled trajectory is serialized into a single sequence that interleaves environment observations and agent outputs across multiple turns. Observations include the current visual feedback (thumbnail overview or returned page images) and the accumulated working-memory summaries from previous turns. Agent outputs include structured \textbf{\texttt{<think>}} content (\emph{analysis/plan/summary} in the first turn; \emph{analysis/relevant\_pages/summary} in later turns) followed by exactly one action tag. During training, the model is conditioned on the entire serialized prefix, but only the agent-generated tokens contribute to the loss. 

\paragraph{\textbf{SFT: Objective}}
Let a serialized trajectory be the token sequence $x_{1:L}$. We define a mask
$m_\ell\in\{0,1\}$ indicating whether token $x_\ell$ belongs to the
agent-generated part (\texttt{<think>} and \texttt{<action>}) or to the
environment observation. The SFT objective is the masked negative
log-likelihood:
\begin{equation}
\mathcal{L}_{\mathrm{SFT}}(\theta)
=
-\sum_{\ell=1}^{L-1}
m_{\ell+1}\,
\log \pi_{\theta}(x_{\ell+1}\mid x_{1:\ell}).
\label{eq:sft_loss_app}
\end{equation}

\paragraph{\textbf{GRPO: Training Data}}
While SFT enables effective imitation, it inherits teacher biases and does not explicitly optimize exploration efficiency under the interaction budget. We therefore further train the agent with GRPO~\cite{guo2025deepseek}, which optimizes expected trajectory-level reward using group-wise sampled rollouts. GRPO training uses only raw dataset-level supervision $(D,Q,y,P_{\mathrm{gt}})$ without intermediate traces. We select 2{,}048 training examples from MP-DocVQA and DUDE that do not overlap with the SFT training set. To ensure a balanced difficulty distribution, we estimate per-example difficulty using the SFT model: for each $(D,Q)$ we run 4 independent rollouts and count the number of successes (ANLS $\ge 0.7$). We then stratify samples into easy/medium/hard buckets and randomly draw them with proportions 10\%/70\%/20\%, respectively.

For each training sample $(D,Q)$, we run the current policy $\pi_{\theta}$ in the same closed-loop environment to sample a group of $G$ complete trajectories $\{T_1,\dots,T_G\}$ (stochastic decoding). Each trajectory terminates when the agent outputs \texttt{<answer>} or reaches the interaction budget. Each sampled trajectory can be represented as a pair $(c_i, a_i)$, where $c_i$ denotes all conditioning context tokens (all observations, including page images and working memory) and $a_i$ denotes the concatenated agent-generated tokens (all \textbf{\texttt{<think>}} and \textbf{\texttt{<action>}} tokens) in that trajectory.

\paragraph{\textbf{GRPO: Reward}} For a trajectory $T$, we compute
$$
R(T)=
w_{\mathrm{a}}R_{\mathrm{a}}(T)+
w_{\mathrm{e}}R_{\mathrm{e}}(T)+
w_{\mathrm{f}}R_{\mathrm{f}}(T).
\label{eq:reward_total_app}
$$
$R_{\mathrm{a}}$ measures answer correctness. For free-form textual answers, we use thresholded \textbf{\emph{Average Normalized Levenshtein Similarity (ANLS)}}:
$$
R_{\mathrm{a}}(T)=
\mathbb{I}[\mathrm{ANLS}(\hat{y},y)\ge\tau]\,
\mathrm{ANLS}(\hat{y},y),
\label{eq:reward_ans_app}
$$
where $\tau=0.5$ and for identifier-like answers we use \textbf{\emph{Exact Match (EM)}}:
$$R_{\mathrm{a}}(T)=\mathbb{I}[\hat{y}=y]$$
For evidence, let $P_{\mathrm{rel}}(T)$ be the union of all pages listed in \textbf{\texttt{<relevant\_pages>}} across turns. We compute a recall-weighted F-score:
$$
R_{\mathrm{e}}(T)=
\frac{(1+\beta^2)\,pr}{\beta^2p+r},
\quad \beta^2=2,
\label{eq:reward_evi_app}
$$
where
$$p=\frac{|P_{\mathrm{rel}}\cap P_{\mathrm{gt}}|}
{|P_{\mathrm{rel}}|+\epsilon}, \quad r=\frac{|P_{\mathrm{rel}}\cap P_{\mathrm{gt}}|}
{|P_{\mathrm{gt}}|+\epsilon}, $$
where $\epsilon$ is a small constant. Finally $R_{\mathrm{f}}$ penalizes binary invalid outputs (unparseable format,
invalid action arguments, or budget violation), with a reward of 1 for valid output and 0 for invalid output.

\paragraph{\textbf{GRPO: Objective}}
GRPO optimizes relative performance within a sampled group. For each group $\{T_i\}_{i=1}^G$, let $R_i=R(T_i)$. We compute the group-normalized advantage: 
\begin{gather*}
A_i = \frac{R_i-\mu}{\sigma+\epsilon} \\
\mu = \frac{1}{G}\sum_{j=1}^G R_j, \quad \sigma = \sqrt{\frac{1}{G}\sum_{j=1}^G (R_j-\mu)^2 }
\end{gather*}
We then update the policy by maximizing the log-likelihood of sampled actions weighted by $A_i$, using a PPO-style clipped objective at the token level. Let $\pi_{\theta_{\mathrm{old}}}$ denote the policy used to sample the group. For each trajectory $i$ and each agent token position $t$, define the ratio
\begin{equation}
\rho_{i,t}(\theta)=
\frac{\pi_\theta(a_{i,t}\mid c_i,a_{i,<t})}
{\pi_{\theta_{\mathrm{old}}}(a_{i,t}\mid c_i,a_{i,<t})}.
\label{eq:ratio_app}
\end{equation}
The GRPO loss is defined as:
\begin{multline}
\mathcal{L}_{\mathrm{GRPO}}(\theta) = -\frac{1}{G}\sum_{i=1}^G \sum_{t=1}^{|a_i|} \min \Big(
\rho_{i,t}(\theta)A_i, \\
\mathrm{clip}\big(\rho_{i,t}(\theta), 1-\epsilon_c, 1+\epsilon_c\big)A_i
\Big),
\label{eq:grpo_loss_app}
\end{multline}
where $\epsilon_c$ is the clip range. Importantly, the loss is applied only on
agent-generated tokens $a_i$; all environment observation tokens are used only
as conditioning context.

\paragraph{Inference: Coarse-to-Fine Evidence Acquisition}
At test time, we use greedy decoding (temperature \(=0\)) and enforce a maximum of \(T\) interaction steps. Starting from the global thumbnail overview, the agent follows a coarse-to-fine strategy: it uses structural cues in \(\tilde{D}\) to propose candidate pages, employs \(\texttt{retrieval\_page}\) with refined queries to localize evidence, uses \(\texttt{fetch\_page}\) for targeted reading and cross-page completion when needed, updates its working memory via summaries, and terminates with \(\texttt{answer}\) once evidence suffices. The essential idea is not to increase context indiscriminately, but to keep the input in a high signal-to-noise regime by actively selecting what to read.

\section{Training and Inference Configuration}
\label{sec:appendix_implementation}

In this section, we provide the comprehensive hyperparameter settings and configuration details for the training and inference of \textbf{Doc-$V^*$}. All experiments were conducted on a computational node equipped with 8 NVIDIA A100 (80GB) GPUs, implemented in PyTorch using BF16 mixed precision to optimize memory efficiency.

\paragraph{Stage I: Supervised Fine-Tuning (SFT)}
The primary goal of the SFT stage is to initialize the agent with stable tool usage capabilities and reasoning behaviors.
\begin{itemize}
    \item \textbf{Data:} We utilize a filtered dataset comprising 9,019 high-quality interaction trajectories.
    \item \textbf{Optimization:} The model is trained for 3 epochs using the AdamW optimizer with a cosine learning rate scheduler. The initial learning rate is set to $3\times10^{-6}$.
    \item \textbf{Loss Masking:} To focus the model's adaptation on reasoning and planning, the loss is computed exclusively on agent-generated tokens (specifically the contents within \texttt{<think>} and \texttt{<action>} blocks), masking out the user instructions and environment observations.
\end{itemize}

\paragraph{Stage II: Group Relative Policy Optimization (GRPO)}
Following SFT, the agent undergoes reinforcement learning alignment to further refine its decision-making logic.
\begin{itemize}
    \item \textbf{Hyperparameters:} We employ a group size of $G=8$ with a sampling temperature of 1.0 to encourage exploration during the generation phase. The training proceeds for 3 epochs with a reduced learning rate of $2\times10^{-6}$.
    \item \textbf{Reward Configuration:} As outlined in the main text, the composite reward function is defined as $R = \omega_{\mathrm{ans}} R_{\mathrm{ans}} + \omega_{\mathrm{evi}} R_{\mathrm{evi}} + \omega_{\mathrm{struct}} R_{\mathrm{struct}}$. The specific coefficients are set to $\omega_{\mathrm{ans}}=0.6$ (Correctness), $\omega_{\mathrm{evi}}=0.3$ (Evidence Recall), and $\omega_{\mathrm{struct}}=0.1$ (Format Validity).
\end{itemize}

\paragraph{Inference Configuration}
During the evaluation phase, to ensure deterministic and reproducible results, we employ greedy decoding (temperature $= 0$). The maximum interaction horizon is fixed at $T=8$ steps, consistent with the constraints applied during the training phase. 

\section{Details of Datasets}
\label{appendix:dataset}

\paragraph{MP-DocVQA~\cite{tito2023hierarchical}} ~ {a multi-page document visual question answering benchmark that focuses on fine-grained information extraction from scanned documents. Questions often require precise localization of textual or visual elements within a document and explicit reasoning over page indices. The dataset emphasizes accurate page navigation and localized evidence grounding.}

\paragraph{DUDE~\cite{van2023document}} ~ {consists of document images paired with questions that demand detailed visual-textual understanding. Compared to MP-DocVQA, DUDE places stronger emphasis on structured layouts such as forms and tables, and requires robust cross-page navigation to retrieve relevant evidence scattered across multiple pages.}

\paragraph{SlideVQA~\cite{tanaka2023slidevqa}} ~ {a document visual question answering dataset focused on understanding presentation slides. It contains slide documents with diverse visual layouts, including figures, charts, bullet lists, and sparsely distributed text. Documents typically span around 20 pages, and the associated questions require complex reasoning over non-linear reading orders and spatial arrangements, rather than relying solely on sequential textual flow.}

\paragraph{LongDocURL~\cite{deng2025longdocurl} } ~ {composed of web-based multi-modal documents with rich structural diversity, such as headings, hyperlinks, images, and embedded tables. With an average document length of approximately 30 pages, the dataset evaluates long-range retrieval and the ability to locate and synthesize information across distant document sections.}

\paragraph{MMLongBench-Doc~\cite{ma2024mmlongbench} } ~ {designed for long-context multi-modal document understanding. Documents in this benchmark are substantially longer, extending up to 468 pages. The dataset poses significant challenges for scalable page selection, efficient navigation, and multi-hop reasoning over large multi-modal contexts.}

\begin{figure}[t]
    \centering
    \includegraphics[width=\linewidth]{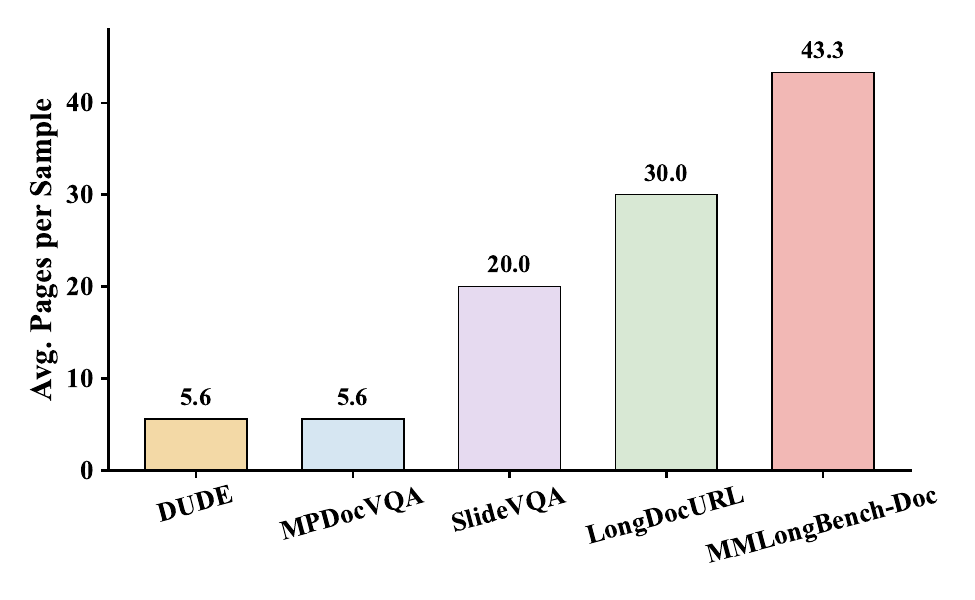} 
    \caption{\textbf{Average document length across datasets.} The figure reports the average number of pages per document for MP-DocVQA, DUDE, SlideVQA, LongDocURL, and MMLongBench-Doc, illustrating the increasing document length and context complexity from standard document QA benchmarks to long-context multi-modal settings.}
    \label{fig:dataset_length}
\end{figure}

\section{Details of Baseline}
\label{sec:appendix_baseline}

\begin{table*}[t!]
\centering
\caption{\textbf{Detailed configurations of Open Source baselines.} ``Retriever'' denotes the model used for page retrieval. ``Param'' refers to the parameter size of the LLM backbone. ``Paradigm'' categorizes methods into End-to-End (\textbf{E2E}), Retrieval-Augmented Generation (\textbf{RAG}), or \textbf{Agent}. The columns under ``Training Dataset'' indicate whether the \textbf{backbone} was fine-tuned ($\checkmark$) on the corresponding benchmark's training set or evaluated in a zero-shot setting ($\times$).}
\label{tab:baseline_details}
\fontsize{9}{11}\selectfont
\setlength\arrayrulewidth{0.6pt}

\providecommand{\venue}[1]{~\textcolor{gray}{\scriptsize\textit{(#1)}}}

\resizebox{1\textwidth}{!}{
\begin{tabular}{l c c c c c c c c}
\toprule
\multirow{2}{*}{\textbf{Method}} & 
\multirow{2}{*}{\textbf{Retriever}} & 
\multirow{2}{*}{\textbf{Backbone}} & 
\multirow{2}{*}{\textbf{Param}} & 
\multirow{2}{*}{\textbf{OCR-Free}} & 
\multirow{2}{*}{\textbf{Paradigm}} & 
\multicolumn{3}{c}{\textbf{Training Dataset}} \\

\cmidrule(lr){7-9} 

 & & & & & & \textbf{DUDE} & \textbf{MPDocVQA} & \textbf{SlideVQA} \\
\midrule

HiVT5 \venue{PR}            & -                  & DiT / T5            & 0.3B & $\times$     & E2E   & $\times$ & $\checkmark$ & $\times$ \\
CREAM \venue{ACM MM'24}     & bge-large          & Pix2Struct / LLaMa2 & 7B   & $\times$     & RAG   & $\checkmark$ & $\checkmark$ & $\times$ \\
mPLUG-DocOwl2 \venue{ACL'25}& -                  & ViT / LLaMa         & 8B   & $\checkmark$ & E2E   & $\checkmark$ & $\checkmark$ & $\times$ \\
M3DocRAG \venue{arXiv'24}   & Colpali            & Qwen2-VL            & 7B   & $\checkmark$ & RAG   & $\times$ & $\times$ & $\times$ \\
VisRAG \venue{ICLR'25}      & VisRAG-Ret         & MiniCPM-V 2.6       & 8B   & $\checkmark$ & RAG   & $\times$ & $\times$ & $\times$ \\
SV-RAG \venue{ICLR'25}      & SV-RAG-InternVL2   & InternVL2           & 4B   & $\checkmark$ & RAG   & $\times$ & $\times$ & $\checkmark$ \\
VDocRAG \venue{CVPR'25}     & VDocRetriever      & Phi3-Vision         & 4B   & $\checkmark$ & RAG   & $\checkmark$ & $\times$ & $\times$ \\
Docopilot \venue{CVPR'25}   & -                  & InternVL2           & 8B   & $\times$     & E2E   & $\checkmark$ & $\checkmark$ & $\times$ \\
DocVLM \venue{CVPR'25}      & -                  & Qwen2-VL            & 7B   & $\times$     & E2E   & $\times$ & $\times$ & $\times$ \\
InternVL3 \venue{arXiv'25}  & -                  & InternViT / Qwen2.5 & 8B   & $\checkmark$ & E2E   & $\times$ & $\times$ & $\times$ \\
VRAG-RL \venue{NeurIPS'25}  & ColQwen2           & Qwen2.5-VL          & 7B   & $\checkmark$ & Agent & $\times$ & $\times$ & $\checkmark$ \\
MoLoRAG \venue{EMNLP'25}    & Colpali+Qwen2.5-VL & Qwen2.5-VL          & 7B   & $\checkmark$ & RAG   & $\times$ & $\times$ & $\times$ \\
CogDoc \venue{arXiv'25}     & -                  & Qwen2.5-VL          & 7B   & $\checkmark$ & Agent & $\checkmark$ & $\times$ & $\checkmark$ \\
URaG \venue{AAAI'26}        & Qwen2.5-VL (Early Layers) & Qwen2.5-VL   & 7B   & $\checkmark$ & RAG   & $\checkmark$ & $\checkmark$ & $\checkmark$ \\
\midrule 
\rowcolor{LLightGray} 
\textbf{Ours}     & Colqwen2.5         & Qwen2.5-VL          & 7B   & $\checkmark$ & Agent & $\checkmark$ & $\checkmark$ & $\times$ \\
\bottomrule
\end{tabular}
}
\end{table*}

This section provides detailed specifications for the open-source baselines compared in our study. Table~\ref{tab:baseline_details}  summarizes their key configurations and training settings, followed by comprehensive descriptions of each method's architecture and paradigm.
\paragraph{HiVT5}
HiVT5~\cite{tito2023hierarchical} proposes a hierarchical multimodal transformer to extend Document VQA to multi-page scenarios, addressing the quadratic complexity of standard attention mechanisms. Relying on an off-the-shelf OCR engine for text and bounding box extraction, it employs a T5-based encoder to process each page independently. The model fuses OCR tokens, layout embeddings, and visual features into learned [PAGE] tokens, which summarize page content conditioned on the query. These summaries are concatenated for the decoder to generate the answer, supported by a module predicting evidence page indices. Training involves a hierarchical layout-aware pre-training task followed by fine-tuning on MP-DocVQA.
\paragraph{CREAM}
CREAM~\cite{zhang2024cream} presents a framework integrating coarse-to-fine retrieval with multimodal efficient tuning to handle token limitations in multi-page documents. It first utilizes an OCR engine to extract and chunk text, followed by a two-stage retrieval process: a coarse ranking via text embedding similarity and a fine-grained re-ranking where an LLM recursively groups chunks to select the top-k candidates. To incorporate visual context, a multi-page vision encoder employs attention pooling to merge features into a unified representation. Based on LLaMA-Adapter V2, the model undergoes multimodal instruction tuning (using LoRA and prefix tuning) to jointly optimize the LLM with retrieved chunks and visual embeddings.
\paragraph{mPLUG-DocOwl2}
mPLUG-DocOwl2~\cite{hu2025mplug2} introduces a modularized Multimodal Large Language Model (MLLM) specialized for OCR-free document understanding. Improving upon the mPLUG-Owl architecture, it employs a visual abstractor to bridge the pre-trained visual encoder and the LLM, directly aligning visual features with textual semantics to eliminate external OCR dependency. The model is optimized via a unified instruction tuning strategy on a diverse document instruction dataset (covering tables, charts, and webpages), enhancing its capability to comprehend fine-grained visual text and complex structures.
\paragraph{M3DocRAG}
M3DocRAG~\cite{cho2024m3docrag} proposes a multimodal Retrieval-Augmented Generation (RAG) framework to overcome the limitations of text-based pipelines in visually rich, open-domain tasks. Diverging from OCR-dependent methods, it adopts an all-multimodal paradigm using a vision-language retriever (e.g., ColPali) to encode page images into visual embeddings. This enables precise retrieval via late interaction mechanisms that preserve layout semantics. The retrieved top-k raw page images are then fed into an MLLM (e.g., Qwen2-VL) for end-to-end question answering. The authors also introduce M3DocVQA, a benchmark requiring cross-document retrieval and multi-hop reasoning.
\paragraph{VisRAG}
VisRAG~\cite{yu2024visrag} presents a vision-based RAG framework that treats document pages purely as images, mitigating information loss from OCR extraction. It employs a dual-encoder architecture (VisRAG-Ret) where queries and document images are encoded into a shared embedding space using position-weighted mean pooling. Generation (VisRAG-Gen) is handled by a generative VLM that synthesizes answers directly from the retrieved visual context. The retriever is fine-tuned via contrastive learning on a mixture of public VQA datasets and synthetic query-document pairs to ensure robust generalization.
\paragraph{SV-RAG}
SV-RAG~\cite{chen2024sv} leverages a single MLLM backbone equipped with two distinct Low-Rank Adaptation (LoRA) adapters to handle both retrieval and generation without external parsers. It employs a retrieval adapter using contextualized late interaction to identify evidence pages, and a QA adapter for answer generation. The adapters are optimized via contrastive learning for retrieval and autoregressive generation for QA, enabling efficient, unified visual retrieval and reasoning within a single model architecture.
\paragraph{VDocRAG}
VDocRAG~\cite{tanaka2025vdocrag} introduces a visual RAG framework designed to process visually rich documents by leveraging visual features directly. It employs a dual-component architecture: VDocRetriever, which retrieves relevant page images using dense token representations, and VDocGenerator, which synthesizes answers from these inputs. To align visual and textual information, the authors utilize self-supervised pre-training tasks that adapt Large Vision-Language Models (LVLMs) for retrieval by compressing visual representations into dense tokens, facilitating open-domain document reasoning.
\paragraph{Docopilot}
Docopilot~\cite{duan2025docopilot} proposes a native multimodal framework that eschews external retrieval in favor of scaling the model's intrinsic context processing. Centered on a "retrieval-free" paradigm, the model ingests entire documents as concatenated high-resolution image sequences. It leverages engineering optimizations like Ring Attention and Liger Kernel to manage long contexts (up to 32k tokens). The capability is supported by "Doc-750K," a large-scale dataset with diverse proxy tasks. Training involves Supervised Fine-Tuning (SFT) with multimodal data-packing, allowing the model to process full document contexts in a single forward pass to resolve long-distance dependencies.
\paragraph{DocVLM}
DocVLM~\cite{nacson2025docvlm} presents a model-agnostic framework to enhance VLMs by efficiently integrating OCR-derived text and layout information. It utilizes an OCR encoder to capture textual and spatial details, compressing them into a compact set of learned queries (typically 64) which are projected into the LLM alongside visual features. This approach preserves the original VLM weights. Training follows a two-stage process: aligning the OCR encoder with the frozen VLM via captioning, followed by fine-tuning on DocVQA datasets, achieving high performance with reduced visual token usage.
\paragraph{InternVL3}
InternVL3~\cite{zhu2025internvl3} is a state-of-the-art multimodal large language model (MLLM) developed by OpenGVLab that advances the field through a native multimodal pre-training paradigm, jointly acquiring visual and linguistic capabilities rather than adapting a text-only backbone. By incorporating variable visual position encoding (V2PE) for extended contexts and advanced post-training techniques like mixed preference optimization, the model achieves superior performance on diverse benchmarks, including MMMU and OCR-related tasks. In this study, InternVL3 is utilized as a strong baseline due to its robust optical character recognition (OCR) and document understanding capabilities, serving as a high-standard reference for evaluating the efficacy of the proposed method in visually rich environments.
\paragraph{VRAG-RL}
VRAG-RL~\cite{wang2025vrag} introduces an agentic framework empowering VLMs with iterative reasoning. It defines a unified action space integrating search queries with fine-grained visual perception actions, specifically predicting coordinates for cropping and zooming into information-dense regions to handle resolution bottlenecks. Operating in a "Thought-Action-Observation" loop, the model generates reasoning chains, executes actions to update observations, and iterates until evidence is gathered. The policy is optimized via Group Relative Policy Optimization (GRPO) with a reward function incentivizing both retrieval precision and answer accuracy.
\paragraph{MoLoRAG}
MoLoRAG~\cite{wu2025molorag} proposes a logic-aware retrieval framework capturing both semantic and logical dependencies. It constructs a document-level "page graph" where edges represent semantic similarities. A lightweight VLM acts as a retrieval engine, traversing this graph by evaluating "logical relevance"—the inferential necessity of a page—alongside semantic alignment. This allows the model to uncover logically connected but semantically distant evidence. The framework supports both a training-free mode and a fine-tuned mode where the engine is optimized on synthesized "question-image-relevance" triplets.
\paragraph{CogDoc}
CogDoc~\cite{xu2025cogdoc} proposes a unified, two-stage cognitive framework mimicking human reading patterns to balance scalability and fidelity. It decomposes reasoning into two phases executed by a single VLM: a "Fast Reading" phase (Localization Mode), scanning the document at low resolution to predict page indices based on structural cues; and a "Focused Thinking" phase (Reasoning Mode), processing localized pages at high resolution for grounded reasoning. To avoid policy conflicts in supervised training, it employs Direct Reinforcement Learning (RL from scratch), enabling the model to autonomously learn to alternate between global scanning and local reasoning.
\paragraph{URaG}
URaG~\cite{shi2025urag} introduces a unified framework integrating retrieval and generation within a single MLLM to handle long documents efficiently. Based on the observation that MLLMs exhibit a "coarse-to-fine" attention pattern, the method inserts a lightweight cross-modal retrieval module into the model's early layers (e.g., layer 6). This module acts as an internal evidence selector, computing relevance via late interaction and retaining only the top-k pages while discarding irrelevant tokens from subsequent layers. This "early-exit" mechanism reduces computational overhead for deeper reasoning layers. Training involves pre-training the retrieval module followed by joint fine-tuning of both components.

\begin{table}[t]
\centering
\caption{\textbf{Impact of K on MMLongBench-Doc.} ``Adaptive'' denotes the document-adaptive setting $K = \min(\lceil N/10\rceil, 4)$, where $N$ is the total number of pages.}
\label{tab:iteration_ablation}
\footnotesize
\setlength{\tabcolsep}{5pt}
\renewcommand{\arraystretch}{1.1}
\begin{tabular}{c c c c c c}
\toprule
\multirow{2}{*}{\textbf{K}} & 
\multirow{2}{*}{\textbf{Avg. Pages}} & 
\multirow{2}{*}{\textbf{Overall}} & 
\multicolumn{3}{c}{\textbf{Breakdown}} \\
\cmidrule(lr){4-6}
& & & SIN & MUL & UNA \\
\midrule
\rowcolor{gray!10}
Adaptive & 5.6 & 42.1 & 54.6 & 23.5 & 45.7 \\
1 & 3.0 & 40.5 & 51.6 & 17.0 & 56.1\\
2 & 4.3 & 39.7 & 54.4 & 20.7 & 39.9 \\
3 & 5.4 & 40.1 & 53.3 & 23.0 & 40.8 \\
4 & 6.5 & 41.1 & 53.3 & 24.0 & 43.5 \\
5 & 8.1 & 41.7 & 52.9 & 23.5 & 48.4 \\
\bottomrule
\end{tabular}
\end{table}


\begin{table}[t]
\centering
\caption{\textbf{Impact of maximum interaction steps on MMLongBench-Doc.}}
\label{tab:k_ablation}
\footnotesize
\setlength{\tabcolsep}{5pt}
\renewcommand{\arraystretch}{1.1}
\begin{tabular}{c c c c c c}
\toprule
\multirow{2}{*}{\textbf{Iteration}} & 
\multirow{2}{*}{\textbf{Avg. Pages}} & 
\multirow{2}{*}{\textbf{Overall}} & 
\multicolumn{3}{c}{\textbf{Breakdown}} \\
\cmidrule(lr){4-6}
& & & SIN & MUL & UNA\\
\midrule
3 & 4.2 & 41.1 & 54.0 & 22.5 & 44.4\\
4 & 4.6 & 41.2 & 54.3 & 23.0 & 43.5 \\
5 & 4.9 & 41.4 & 54.3 & 23.4 & 43.5 \\
6 & 5.2 & 41.5 & 54.4 & 24.0 & 44.0 \\
\rowcolor{gray!10}
7 & 5.6 & 42.1 & 54.6 & 23.5 & 45.7 \\
8 & 5.8 & 41.4 & 54.3 & 24.0 & 44.0 \\
9 & 6.0 & 41.5 & 54.8 & 24.0 & 44.0 \\
10 & 6.2 & 41.5 & 54.8 & 24.0 & 44.0 \\
\bottomrule
\end{tabular}
\end{table}

\section{Robustness of Iteration \& K}
\label{appendix:robustness}

We investigate how the number of interaction turns (iterations) affects the agent's performance on the \textbf{MMLongBench-Doc} dataset. As the agent operates in a recursive ``Observe-Think-Act'' loop, the number of steps determines the depth of exploration.
As shown in Table~\ref{tab:iteration_ablation}, performance improves consistently as the maximum iteration limit increases from 3 to 7. The model achieves peak performance at \textbf{7 iterations} with an overall accuracy of \textbf{42.1\%}. This suggests that for complex long-document tasks, the agent requires approximately 5--7 steps to effectively locate evidence and synthesize answers. Beyond 7 iterations, the performance plateaus and slightly fluctuates, indicating that the agent has converged and further exploration yields diminishing returns.

We further analyze the effect of the page selection budget \(K\) on \textbf{MMLongBench-Doc}, as reported in Table~\ref{tab:k_ablation}. Overall performance exhibits a clear non-monotonic trend with respect to \(K\).
When \(K\) is small (e.g., \(K=1\) or \(2\)), the agent is restricted to a limited number of pages, leading to insufficient evidence coverage and degraded overall accuracy. As \(K\) increases, performance improves steadily and reaches its peak under the \textbf{Adaptive} setting, where \(K=\min(\lceil N/10\rceil,4)\). This adaptive strategy achieves the best overall accuracy of \textbf{42.1\%} while maintaining a moderate average page count of 5.6.

Further increasing \(K\) beyond the adaptive range does not result in consistent gains. Although larger \(K\) values introduce more pages, the additional context also brings redundant or irrelevant information, which weakens evidence aggregation and slightly hurts performance. This effect is particularly evident in the SIN and MUL subsets, where accuracy saturates or fluctuates as \(K\) grows. These results indicate that effective long-document reasoning depends on selecting a well-calibrated number of pages rather than aggressively expanding the context. The adaptive strategy strikes a favorable balance between evidence sufficiency and noise control, highlighting the importance of dynamic, document-aware page budgeting.

\section{Case Study}
\label{sec:case_study}

\begin{figure*}[]
    \centering
    \includegraphics[width=0.95\linewidth]{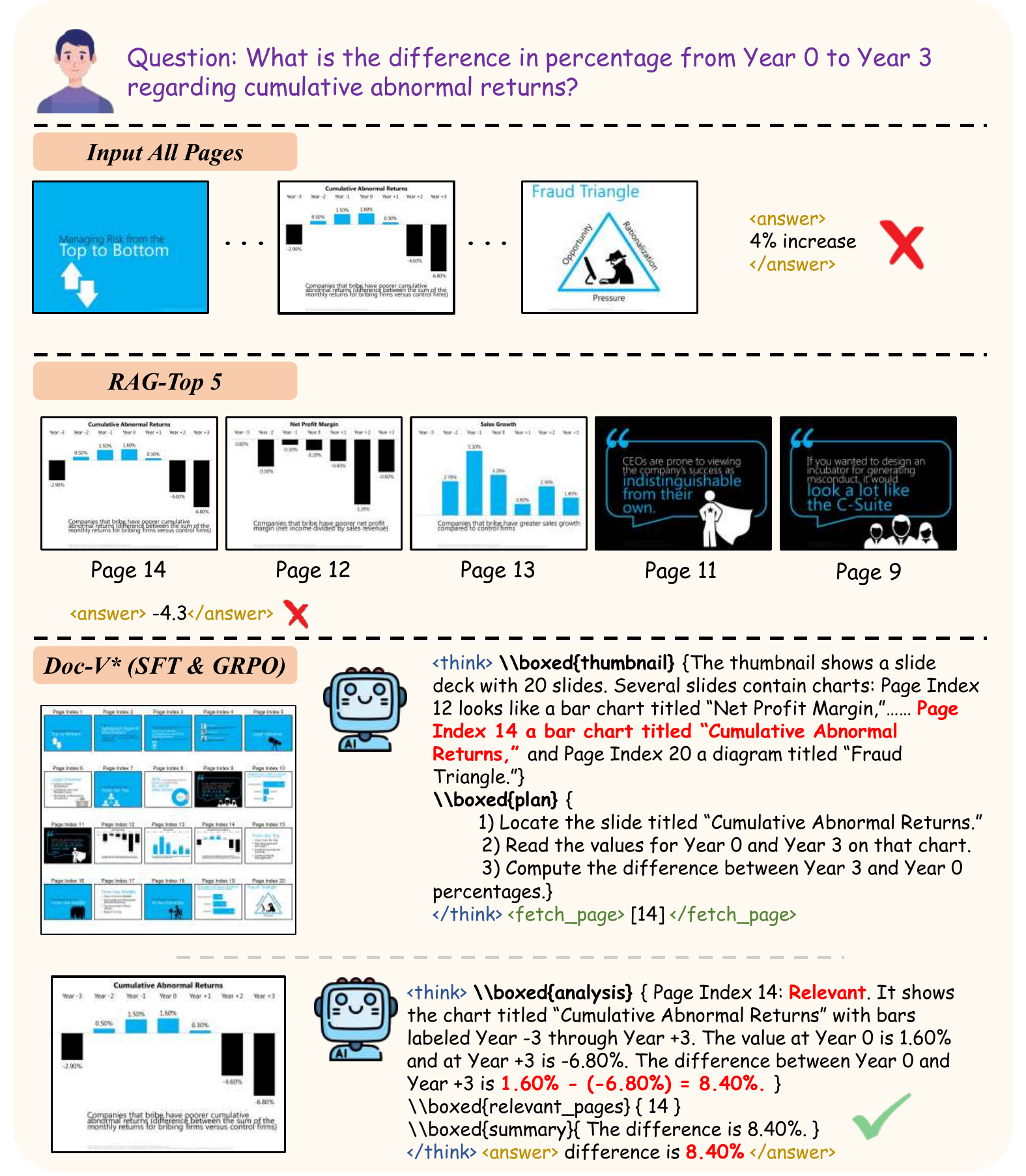} 
    \caption{Case 1 in SlideVQA between different methods}
    \label{fig:slidevqa_case1}
\end{figure*}

\begin{figure*}[]
    \centering
    \includegraphics[width=0.95\linewidth]{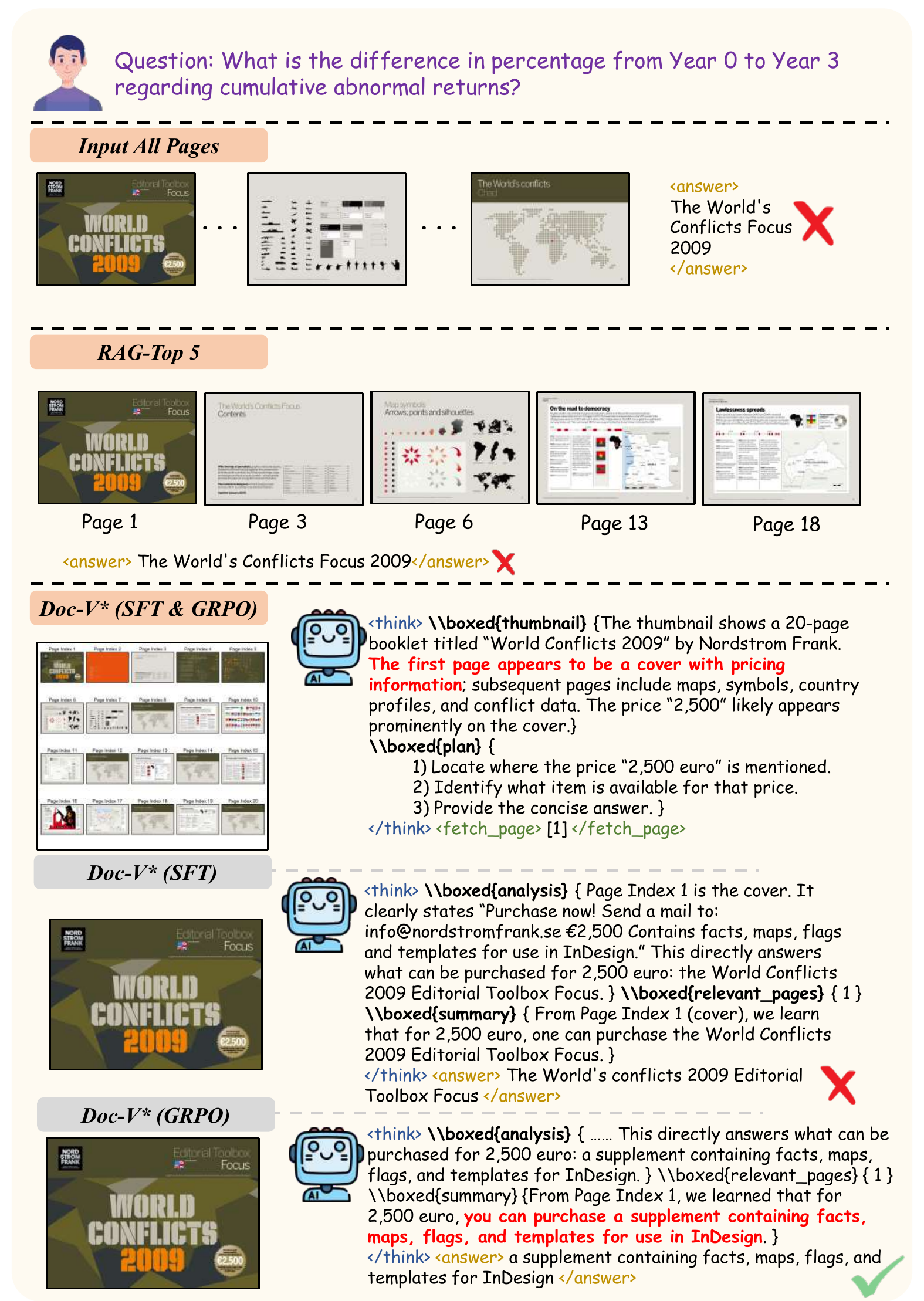} 
    \caption{Case 2 in SlideVQA between different methods}
    \label{fig:slidevqa_case2}
\end{figure*}

\begin{figure*}[]
    \centering
    \includegraphics[width=0.95\linewidth]{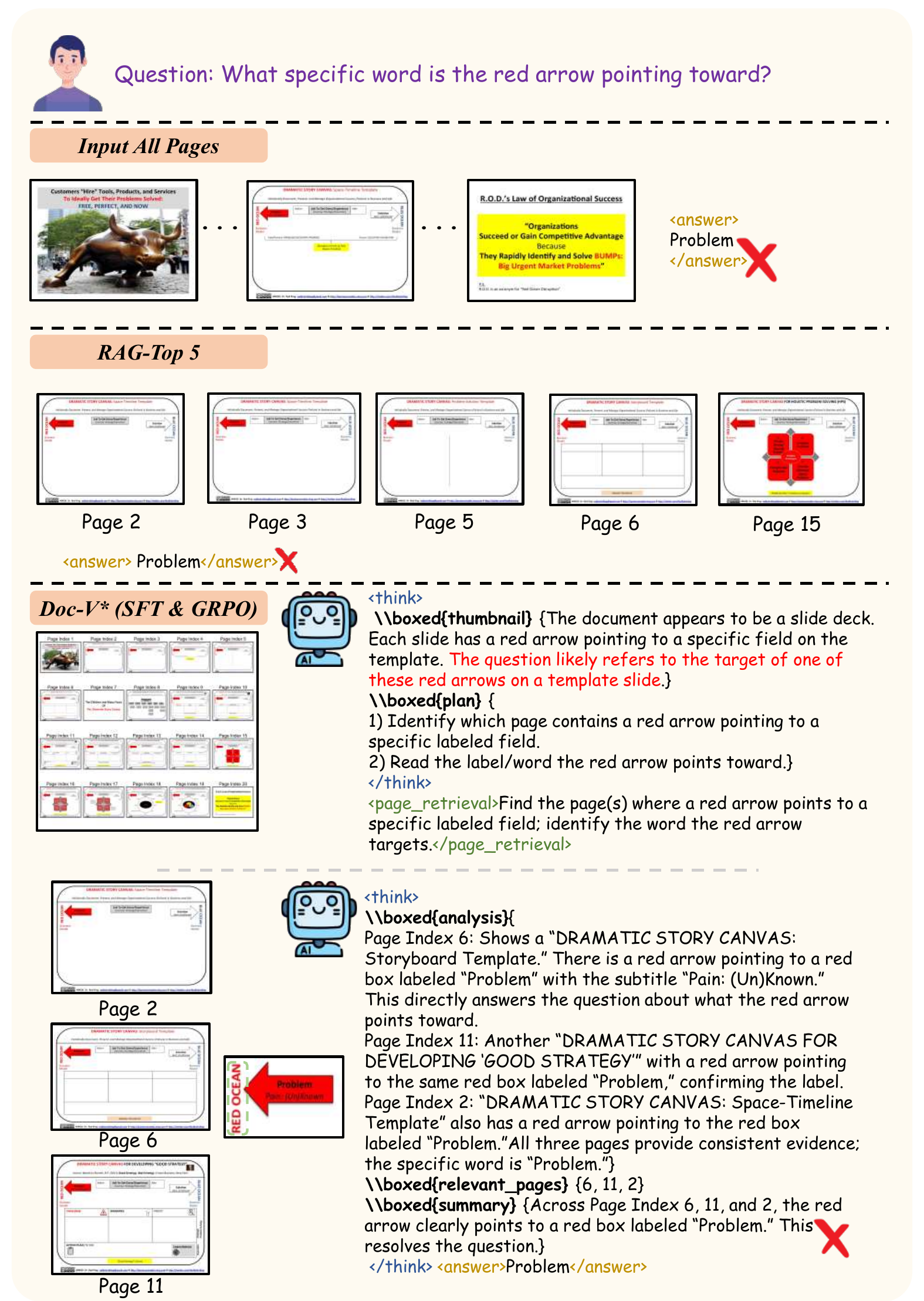} 
    \caption{Case 3 in SlideVQA between different methods}
    \label{fig:slidevqa_case3}
\end{figure*}

See Fiugre~\ref{fig:slidevqa_case1},~\ref{fig:slidevqa_case2},~\ref{fig:slidevqa_case3}

\end{document}